\documentclass{article}

\usepackage[final]{neurips_2024}
\usepackage{times}
\usepackage{latexsym}
\usepackage{multirow}
\usepackage{subfig}
\usepackage{graphicx}
\usepackage{url}
\usepackage{listings}
\usepackage{xcolor}
\usepackage[flushleft]{threeparttable}

\usepackage[utf8]{inputenc} 
\usepackage[T1]{fontenc}    
\usepackage{hyperref}       
\usepackage{url}            
\usepackage{booktabs}       
\usepackage{amsfonts}       
\usepackage{nicefrac}       
\usepackage{microtype}      
\usepackage{xcolor}         

\title{Illusory VQA: Benchmarking and Enhancing Multimodal Models on Visual Illusions}

\author{%
  Mohammadmostafa Rostamkhani \\
  Department of Computer Engineering\\
  IUST\\
  \texttt{mo\_rostamkhani97@comp.iust.ac.ir} \\
  \And
  Baktash Ansari \\
  Department of Computer Engineering \\
  IUST \\
  \texttt{baktash\_ansari@comp.iust.ac.ir} \\
  \AND
  Hoorieh Sabzevari \\
  Department of Computer Engineering \\
  IUST \\
  \texttt{h\_sabzevari@elec.iust.ac.ir} \\
  \And
  Farzan Rahmani \\
  Department of Computer Engineering \\
  IUST \\
  \texttt{farzan\_rahmani@comp.iust.ac.ir} \\
  \And
  Sauleh Eetemadi \\
  Department of Computer Engineering \\
  IUST \\
  \texttt{sauleh@iust.ac.ir} \\
}

\definecolor{codegreen}{rgb}{0,0.6,0}
\definecolor{codegray}{rgb}{0.5,0.5,0.5}
\definecolor{codepurple}{rgb}{0.58,0,0.82}
\definecolor{backcolour}{rgb}{0.95,0.95, 0.95}

\lstdefinestyle{mystyle}{
    backgroundcolor=\color{backcolour},   
    commentstyle=\color{codegreen},
    keywordstyle=\color{magenta},
    numberstyle=\tiny\color{codegray},
    stringstyle=\color{codepurple},
    basicstyle=\ttfamily\footnotesize,
    breakatwhitespace=false,         
    breaklines=true,                 
    captionpos=b,                    
    keepspaces=true,                 
    numbers=left,                    
    numbersep=5pt,                  
    showspaces=false,                
    showstringspaces=false,
    showtabs=false,                  
    tabsize=2
}

\begin{document}

\maketitle

\begin{abstract}
  In recent years, Visual Question Answering (VQA) has made significant strides, particularly with the advent of multimodal models that integrate vision and language understanding. However, existing VQA datasets often overlook the complexities introduced by image illusions, which pose unique challenges for both human perception and model interpretation. In this study, we introduce a novel task called Illusory VQA, along with four specialized datasets: IllusionMNIST, IllusionFashionMNIST, IllusionAnimals, and IllusionChar. These datasets are designed to evaluate the performance of state-of-the-art multimodal models in recognizing and interpreting visual illusions. We assess the zero-shot performance of various models, fine-tune selected models on our datasets, and propose a simple yet effective solution for illusion detection using Gaussian and blur low-pass filters. We show that this method increases the performance of models significantly and in the case of BLIP-2 on IllusionAnimals without any fine-tuning, it outperforms humans. Our findings highlight the disparity between human and model perception of illusions and demonstrate that fine-tuning and specific preprocessing techniques can significantly enhance model robustness. This work contributes to the development of more human-like visual understanding in multimodal models and suggests future directions for adapting filters using learnable parameters.
\end{abstract}

\section{Introduction}
In recent years, there has been significant progress in the field of Visual Question Answering (VQA) and multimodal models, which involves answering questions about images using both vision and language understanding. VQA requires models to interpret visual content and comprehend natural language in order to provide accurate answers. Most existing VQA datasets focus on traditional image understanding and do not consider the challenges posed by novel image illusions.\\
\begin{figure}[h]
 \centering
  \subfloat[Raw image without illusion]{\includegraphics[width=.48\linewidth]{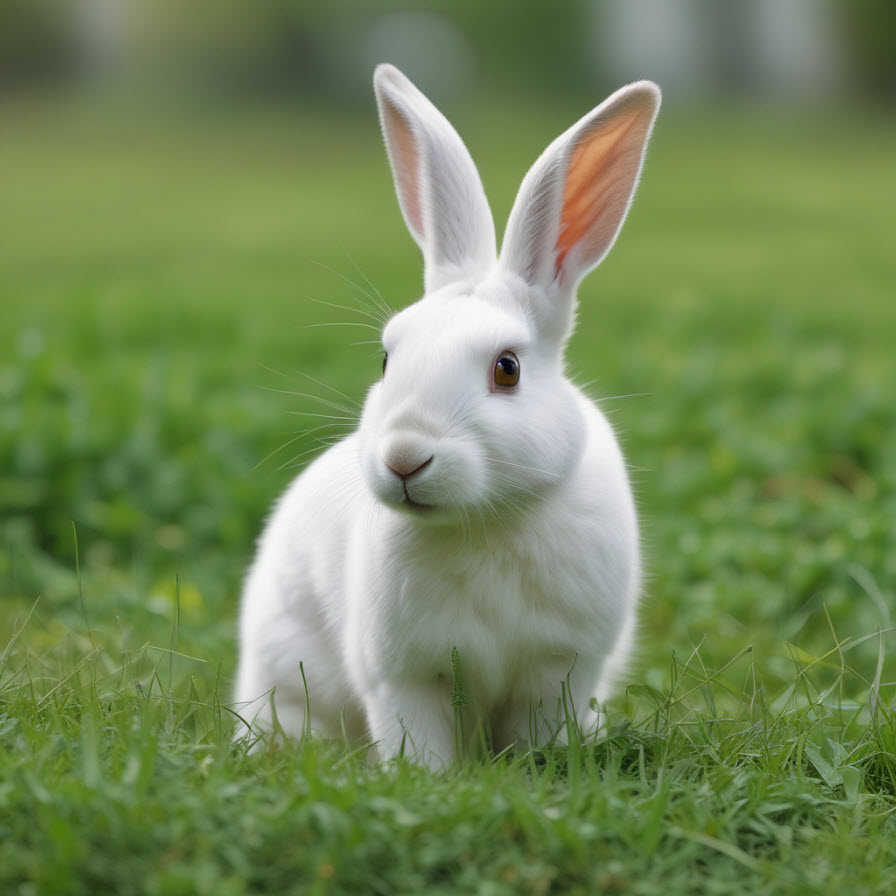}}
  \subfloat[Image illusion]{\includegraphics[width=.48\linewidth]{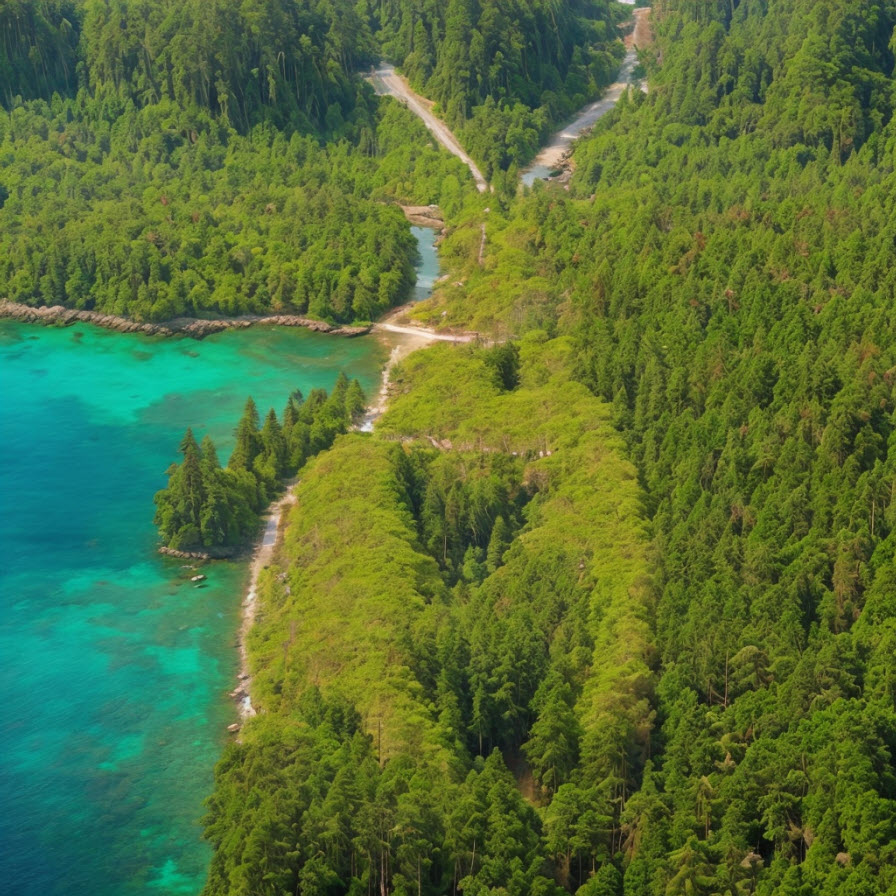}}
  \caption{Illustration of illusion in image}
  \label{fig: illusion illustration}
\end{figure}

\begin{figure*}[h]
\centering
    \includegraphics[width=0.98\linewidth]{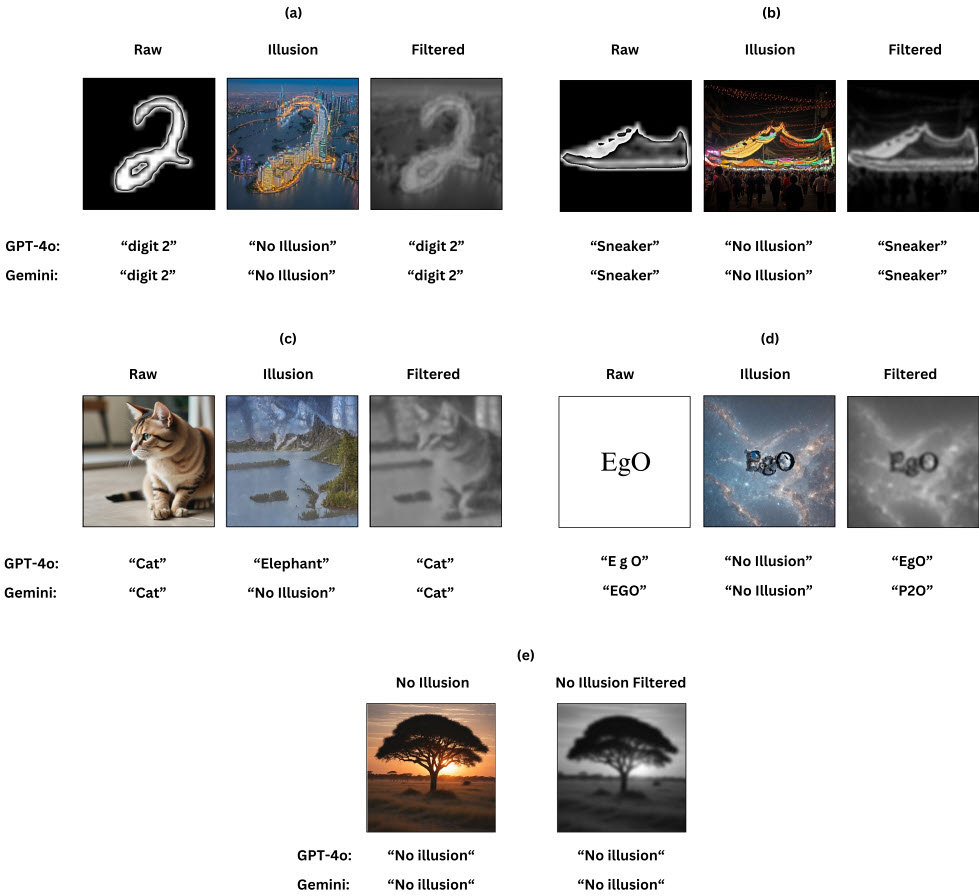}
\caption{Examples of answers of GPT-4o and Gemini to raw images, illusory images, and applied filter one. Through the process of image minification, the human eye becomes adept at uncovering the hidden illusions within. (a): An example of IllusionMNIST, (b): An example of IllusionFashionMNIST, (c): An example of IllusionAnimals, (d): An example of IllusionChar, (e): An example of 'No illusion' class}
\label{fig: dataset example}
\end{figure*}

Image illusion refers to the phenomenon where an image initially portrays one thing, yet upon closer examination and a slight adjustment of focus, it reveals an entirely different representation. An example of this can be found in Figure \ref{fig: illusion illustration}. The phenomenon we are discussing is called "pareidolia," a psychological effect where people perceive familiar patterns or images (such as faces, animals, or objects) in random or unrelated stimuli. In this instance, you initially see one image, but upon closer inspection, your brain detects another image hidden within the scene. Our brains are naturally wired to recognize patterns, which explains why we often identify familiar shapes in clouds, rock formations, or in images like the one presented here. This is also referred to as a "visual illusion" or "perceptual illusion."
In this case, the image consists of an arrangement of shapes and colors that your brain interprets in multiple ways—first as a landscape, and then, upon shifting your perception, as the shape of a rabbit. This type of illusion occurs because the brain attempts to make sense of ambiguous or complex visuals by aligning them with familiar patterns. Although the image itself remains unchanged, your perception shifts, giving the impression of seeing something not intentionally designed into the image. Essentially, illusions like this illustrate how our brains actively interpret and construct reality based on visual input. In this work, we use the terms "pareidolia" and "illusion" interchangeably.\\
The inclusion of illusion images in the dataset is important because it opens up possibilities for applications in steganography and potentially bypassing ethical rules. These types of images for multimodal models act as brainteasers for language models. While the image may contain certain elements, one needs to view it from a different perspective to perceive the illusion within it. It is widely recognized that the human perceptual systems are prone to visual illusions, which can be described as “consistent and persistent discrepancies between a physical state of affairs and its representation in consciousness” \cite{RHDay}. In recent years, there has been a growing interest in model-perceived visual illusions, inspired by the fascinating phenomena observed in human perception \cite{ConvolutionalNeuralNetworksCanBeDeceivedbyVisualIllusions,biologyapproachtoDNN, color-visual-illusions, deepneuralnetworkexhibitsillusion}. These previous studies have primarily focused on vision, exploring how computer vision models can replicate the effects of illusions by examining the internal representations and comparing them to the perceptual shifts experienced by humans. In spite of the emergence of multimodal models such as CLIP \cite{CLIP}, BLIP \cite{BLIP}, BLIP-2 \cite{BLIP-2}, Kosmos-2 \cite{kosmos-2}, LLaVA \cite{llava}, MiniGPT-V2 \cite{minigpt-4, minigpt4-v2}, Gemini \cite{gemini}, and GPT-4V(ision) \cite{GPT-4V}, \textbf{a significant disparity remains between the way humans perceive and interpret images and how these models perceive them}. We can use these illusory datasets to make VLMs behave more like humans. We suggest that this approach can also enhance their robustness in OCR capabilities.\\

In this study, our main contributions are as follows.
\begin{enumerate}
    \item Introduction of new task called Illusory VQA
    \item Introduction of IllusionMNIST, IllusionFashionMNIST, IllusionAnimals, and IllusionChar Datasets \footnote{All code and results are accessible at \url{https://github.com/IllusoryVQA/IllusoryVQA}, and the datasets can be found at \url{https://huggingface.co/VQA-Illusion}.
}: We develop these datasets specifically for benchmarking multimodal models. These datasets include separate train and test sets, providing a comprehensive evaluation framework. Some examples from these datasets can be found in Figure \ref{fig: dataset example}. 
    \item Evaluation of Zero-Shot Performance: We assess the zero-shot performance of state-of-the-art multimodal models on the aforementioned datasets. This evaluation helps us understand how well these models can generalize and interpret the illusions present in the images.
    \item Fine-Tuning and Re-evaluation: We further enhance the evaluation by fine-tuning some of the multimodal models on the training set and then re-evaluating their performance on the test set. This process allows us to measure the effectiveness of fine-tuning in improving the models' ability to detect illusions.
    \item Proposal of an Effective Solution for Illusion Detection: Additionally, we propose a simple yet effective solution for detecting illusions in images. This solution aims to enhance the models' capability to identify and differentiate between real and illusory elements in the datasets. Our approach involves the application of a fixed Gaussian and blur filter to illusory images. You can find the details of the filter we used in the appendix \ref{appendix: filter}.
\end{enumerate}

\section{Related Work}
\cite{image-distortion} explores illusory contour perception in deep learning models by creating illusory contour datasets through image distortion techniques based on the abutting grating illusion. It also aims to systematically generate illusory contour samples, construct illusory contour versions of datasets like MNIST, and test these illusions in various deep learning models from TorchVision. \cite{optical-illusion-images-dataset} creates an optical illusion images dataset consisting of 6725 illusion images from different sources and a smaller dataset of 500 hand-picked images. \cite{synthesis-of-visual-illusions} propose a framework for synthesizing visual illusions using deep generative models to understand the differences between vision models and human perception. It also aims to create novel visual illusions by optimizing GANs to produce illusions that have a maximum effect on a given vision model. \cite{color-visual-illusions} explains visual illusions, particularly lightness and color visual illusions, through the likelihood of patches in natural images. It introduces a computational model that computes the likelihood of patches based on a large dataset, providing a data-driven explanation for visual illusions. It also highlights the role of retinal image statistics in shaping visual perception and cognition, emphasizing the importance of studying visual input. \cite{whoops} introduces the WHOOPS dataset and benchmark, focusing on visual commonsense reasoning by presenting unconventional and commonsense-defying images created synthetically. The dataset consists of 500 purposefully unconventional images challenging AI models' visual commonsense reasoning abilities. Tasks include image captioning, cross-modal matching, visual question answering, and a challenging explanation generation task. The dataset aims to inspire the development of AI models with stronger visual commonsense reasoning abilities. \cite{hallusionbench-illusion} introduces "HALLUSIONBENCH",  a benchmark designed to evaluate image-context reasoning LVLMs and specifically designed to test the capabilities of LVLMs in handling visual illusions and language hallucinations. This benchmark challenges advanced LVLMs like GPT-4V, Gemini Pro Vision, Claude 3, and LLaVA1.5 by emphasizing nuanced understanding and interpretation of visual data. It Identifies failure modes related to visual illusion and language hallucination and provides insights into the challenges posed by visual illusion and hallucination in large vision-language models, paving the way for potential improvements in these models. Also \cite{HALLUSIONBENCH-see-think} provides an in-depth analysis of the failures of state-of-the-art LVLMs like GPT-4V and LLaVA-1.5 on HALLUSIONBENCH. It identifies and categorizes the types of mistakes these models make, attributing them to either language hallucination or visual illusion. \cite{grounding-visual-illusion} investigates how VLMs align with human visual perception under the influence of visual illusions. The study explores whether models interpret visual information similarly to humans when both are subjected to illusions. They introduce the Grounding Visual Illusion in Language (GVIL) benchmark. This benchmark comprises four subtasks (Same-Difference Question Answering, Referential Question Answering, Attribute Question Answering, and Referential Localization) to assess the alignment between human and model interpretations under visual illusions. \cite{illusionvqa} evaluates the capability of VLMs in understanding and interpreting optical illusions. The study introduces a new dataset, IllusionVQA, which consists of challenging optical illusions designed to test VLMs in two specific tasks: comprehension and soft localization. \cite{diffusionillusionshidingimages} introduces the Diffusion Illusions framework, which utilizes a frozen text-to-image diffusion model to automatically generate various optical illusions, including flip illusions, rotation overlay illusions, and hidden overlay illusions. The approach leverages score distillation and dream target losses to optimize prime images, demonstrating successful real-world fabrication and expanding the possibilities for creating and using optical illusions. \cite{visualanagramsgeneratingmultiview} presents a novel method for generating multi-view optical illusions using diffusion models, creating images that reveal different interpretations when subjected to transformations like flips, rotations, or pixel rearrangements. The approach leverages the intrinsic capabilities of diffusion models to produce these illusions without the need for explicit perceptual models. \cite{bistable} evaluates how various vision-language models interpret bistable images, revealing that most models exhibit strong biases toward one interpretation, largely influenced by language priors rather than visual data. The findings suggest that current models struggle to match human-like perception of ambiguous images, indicating a need for further research in handling visual ambiguity. In these types of images viewers can only see one interpretation at a time. \cite{codis} introduces CODIS, a novel benchmark designed to evaluate the context-dependent visual comprehension of Multimodal Large Language Models (MLLMs), highlighting the models' current limitations in leveraging contextual information to disambiguate images.

\section{Task and Dataset}
In this section, we propose a task definition for Illusory VQA and discuss some challenges associated with this task, as well as the process of creating our datasets.

\subsection{Task Definition}
In the Illusory VQA task, the goal is to analyze illusory images using a Vision-Language Model (VLM). These images, denoted as \textbf{\textit{Illusory Images (II)}}, depict both a \textbf{\textit{Real Concept (RC)}} and \textbf{\textit{potentially an Illusory Concept (IC)}}. Given an \textbf{\textit{Illusory Image (II)}} and a \textbf{\textit{question (Q)}} about the image, the VLM must first detect if there is an illusory concept present and then provide an \textbf{\textit{answer (A)}} specifically related to the illusory concept in the image. 

In this study, we focus exclusively on the illusory concept, ensuring that the model can accurately detect and respond to questions about it. This means that the primary objective is to evaluate the model's ability to identify the illusory concept within the image and provide a correct answer related to that concept. You can find an example of the task definition in the appendix \ref{appendix: Task Examples}.

\subsection{Challenges of the Task}
One of the primary challenges of the Illusory VQA task is accurately detecting the presence of an illusory concept (\textit{IC}) in the image. Despite the presence of the real concept (\textit{RC}) in the image, the model must still answer questions pertaining to the illusory concept. For example, in classification tasks, there may be an illusion of one class in the picture, but instances of another class may also be present. In such cases, the model must correctly detect the illusory concept and provide accurate answers to questions about it.\\
By addressing these challenges, the Illusory VQA task aims to enhance our understanding of how VLMs perceive and interpret illusory images, as well as their ability to reason about illusory concepts within the context of a given question.

\subsection{Data Generation}
\begin{figure*}[h]
\centering
    \includegraphics[width=0.98\linewidth]{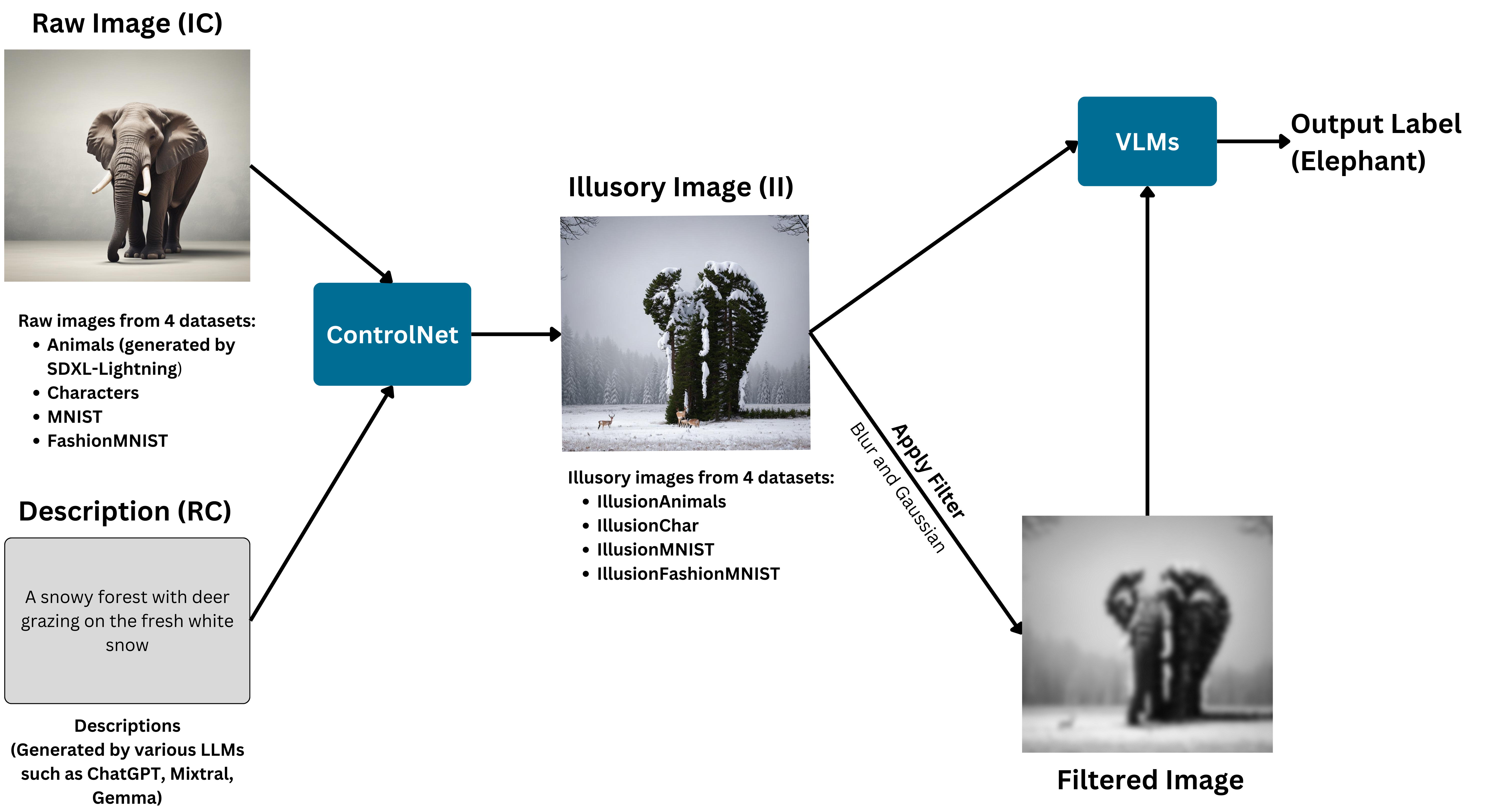}
\caption{Demonstraion of our pipeline for generating and evaluating datasets}
\label{fig: data generation}
\end{figure*}

To create and collect illusory images, we initially generate 1027 descriptions of scenes for transferring images to new space, using various LLMs such as ChatGPT, Gemini, Mixtral-8x7B-Instruct-v0.1 \cite{mixtral}, and Gemma-1.1-7b-it \cite{gemma} to ensure diversity of descriptions. We use these descriptions to generate the real concept of our images (\textit{RC}). We ensured that the number of descriptions generated from each LLM was approximately equal. After collecting our raw images (\textit{IC}), we combine them with the descriptions (\textit{RC}) generated by the LLMs. Using a variant of ControlNet model \cite{controlnet}, we utilized this combined data to generate illusory images (\textit{II}). The demonstration of the whole process is shown in Figure \ref{fig: data generation}. While these datasets are generated synthetically, we believe they hold significant value. The quality of the generated images has been validated by human reviewers, which enhances their reliability. Additionally, these datasets can assist multimodal models in understanding images more effectively, similar to human perception. In order to ensure the safety and appropriateness of our datasets, we took several precautionary measures. Initially, we assumed that our original datasets, namely MNIST and Fashion-MNIST, as well as the animal dataset generated by SDXL Lightning, did not contain any offensive content. Additionally, we confirmed that the Char dataset also does not include any offensive content.
To further validate the absence of offensive content, we employed GPT-3.5 to examine and verify the descriptions generated for the datasets. This step was crucial in ensuring that our descriptions were free from any offensive material. Based on these measures, we initially believed that our datasets were free from offensive content.
However, to provide further assurance, we conducted a sampling process. We randomly selected 10\% of the data from each class in our datasets and presented them to human annotators. Their task was to identify any offensive content within the samples. This process allowed us to determine whether any offensive content was present in our datasets.
Surprisingly, during this evaluation, we discovered that a portion of the images in our datasets did indeed contain offensive content. To address this, we employed NSFW (Not Safe for Work) detector models on our dataset images. By running these models, we were able to identify and filter out the images that contained offensive content.
In our study, we conducted experiments using a dataset consisting of content that has the potential to be offensive. For the purpose of publishing our datasets, we have excluded the data that was flagged by the NSFW detector models. The datasets that we are making available for public use have undergone this additional filtering process to ensure that they are free from offensive content.

\begin{table*}[]
\caption{Size of provided datasets}
\label{tab: dataset statistics}
\centering
\resizebox{0.7\linewidth}{!}{
\begin{tabular}{|c|c|c|}
\hline
\textbf{Dataset} & \textbf{\# of training samples} & \textbf{\# of test samples} \\ \hline
\textbf{IllusionMNIST} & 3960 & 1219 \\ \hline
\textbf{IllusionFashionMNIST} & 3300 & 1267 \\ \hline
\textbf{IllusionAnimals} & 3300 & 1100 \\ \hline
\textbf{IllusionChar} & 9900 & 3300 \\ \hline
\end{tabular}
}
\end{table*}

\subsubsection{IllusionMNIST}
To create IllusionMNIST, we start by selecting MNIST\footnote{The MNIST dataset consists of 10 classes representing digits from 0 to 9.} \cite{mnist} as our source for raw images. From there, we sample images and resize them to a resolution of 512 pixels by 512 pixels. These images were then combined with descriptions generated by LLMs, and IllusionMNIST was formed using a variant of ControlNet. In order to further challenge the models, we introduced an additional class called 'No illusion' to the original set of classes, as we can see in Figure \ref{fig: dataset example}. This class enables the models to detect instances where no illusion images were present in the picture. To ensure fairness in the dataset, we ensure that an equal number of samples are included for the 'No illusion' class. Ultimately, our dataset consists of  3960 training samples and 1219 test samples. The statistics of our datasets are shown in Table \ref{tab: dataset statistics}. For more information, please refer to Appendix \ref{appendix:pie_chart}.

\subsubsection{IllusionFashionMNIST}
For the creation of IllusionFashionMNIST, we opt to use Fashion-MNIST\footnote{The Fashion-MNIST dataset contains 10 classes representing items such as 'T-shirt/top', 'Trouser', 'Pullover', 'Dress', 'Coat', 'Sandal', 'Shirt', 'Sneaker', 'Bag', and 'Ankle boot'.} \cite{fashionmnist} as the source for our raw images. Similar to IllusionMNIST, we sample images from the dataset and resize them to a resolution of 512 pixels by 512 pixels. These images were then combined with descriptions generated by LLMs, and IllusionFashionMNIST was formed using ControlNet. Just like in the previous case, we include an additional class called 'No illusion' to provide a challenge to the models. In the end, we obtain a dataset of 3300 training samples and 1267 test samples.

\subsubsection{IllusionAnimals}
To generate IllusionAnimals, we create a dataset\footnote{The dataset created for IllusionAnimals comprises 10 classes representing animals such as 'cat', 'dog', 'pigeon', 'butterfly', 'elephant', 'horse', 'deer', 'snake', 'fish', and 'rooster'.} consisting of raw animal images using SDXL-Lightning \cite{sdxllightning}. These images were then combined with descriptions generated by LLMs and transformed into IllusionAnimals using ControlNet. Once again, we add an extra class called 'No illusion' to push the models' capabilities. The final dataset for IllusionAnimals comprises 3300 training samples and 1100 test samples.

\subsubsection{IllusionChar}
To explore problems beyond classification, we develop a dataset called IllusionChar, which centers around reading characters in pictures. The creation of the IllusionChar dataset involves a two-step process. Initially, we generate images consisting of a sequence of characters, ranging from 3 to 5 characters per image. Subsequently, we apply an illusion effect to these characters using ControlNet. Just like the other datasets, we include some images with 'No illusion' to provide diversity. As a result, the IllusionChar dataset contains 9900 training samples and 3300 test samples.\\
\\
\textbf{Human Validation and Evaluation } To assess the quality of the generated dataset, we conduct a human evaluation involving a panel of participants. For each dataset, we select three individuals to participate.
In the evaluation and validation process, we randomly sample 10\% of each dataset (10\% of each class for classification datasets) and present the images to the participants. They were asked to identify the class or category they perceived in each picture. To ensure the validity of the generated images, we also disclose the true label after they choose their label to the participants and ask them to confirm if they perceive the indicated class or not.\\
please consult the Appendix \ref{appendix:human} for more details.

\section{Experimental Setup}

We conduct evaluations on various models for IllusionMNIST, IllusionFashionMNIST, IllusionAnimals, and IllusionChar. Specifically, we assess the performance of GPT-4o, Gemini 1.0 Pro Vision, LLaVA-1.5-7B, Kosmos-2-patch14-224, MiniGPT-V2 based on Llama2 Chat 7B (after stage-3 pretrained weights), CLIP-ViT-base-batch32, BLIP-Large, and BLIP-2 on IllusionMNIST, IllusionFashionMNIST, and IllusionAnimals. We use Hugging Face and LAVIS \cite{lavis} for implementing these models. For IllusionChar, we focused on evaluating GPT-4o and Gemini 1.0 Pro Vision.\\
For the fine-tuning phase on IllusionMNIST, IllusionFashionMNIST, and IllusionAnimals, we utilize LLaVA, CLIP, BLIP, and BLIP-2. However, due to hardware limitations, we employed LoRA \cite{lora} for the LLaVA model. For detailed information about implementation, please refer to Appendix \ref{appendix:hyperparameters}.

\begin{table*}[]
\caption{Zero-shot performance of different models on different datasets: The term 'Raw' refers to raw images without any illusions. 'Illusion' refers to illusory images, while 'Filtered' indicates illusory images that have been processed with our filter.}
\label{tab: zero-shot}
\centering
\resizebox{\linewidth}{!}{
\begin{threeparttable}
\begin{tabular}{cc|cccc|cccc|cccc|}
\cline{3-14}
 &  & \multicolumn{4}{c|}{\textbf{IllusionMNIST}} & \multicolumn{4}{c|}{\textbf{IllusionFashionMNIST}} & \multicolumn{4}{c|}{\textbf{IllusionAnimals}} \\ \cline{3-14} 
 &  & \textbf{Accuracy} & \textbf{Precision} & \textbf{Recall} & \textbf{F1} & \textbf{Accuracy} & \textbf{Precision} & \textbf{Recall} & \textbf{F1} & \textbf{Accuracy} & \textbf{Precision} & \textbf{Recall} & \textbf{F1} \\ \hline
\multicolumn{1}{|c|}{\multirow{3}{*}{\textbf{GPT-4o}\tnote{*}}} & \textbf{Raw} & \textbf{89.88} & \textbf{90.26} & \textbf{85.89} & \textbf{86.99} & 67.12 & \textbf{71.05} & \textbf{67.89} & \textbf{65.93} & \textbf{100} & \textbf{100} & \textbf{100} & \textbf{100} \\
\multicolumn{1}{|c|}{} & \textbf{Illusion} & \textbf{37.47} & 44.94 & \textbf{37.40} & \textbf{36.95} & \textbf{16.67} & 27.76 & \textbf{16.61} & \textbf{12.72} & 34.89 & 35.89 & 27.45 & 27.08 \\
\multicolumn{1}{|c|}{} & \textbf{Filtered} & \textbf{67.97} & 70.39 & \textbf{67.99} & \textbf{68.32} & \textbf{48.73} & 48.76 & 44.73 & 41.58 & 83.99 & 72.60 & 71.08 & 70.39 \\ \hline
\multicolumn{1}{|c|}{\multirow{3}{*}{\textbf{Gemini}}} & \textbf{Raw} & 89.36 & 76.23 & 74.27 & 74.26 & \textbf{67.27} & 40.56 & 37.43 & 36.74 & 87.20 & 76.92 & 67.08 & 68.68 \\
\multicolumn{1}{|c|}{} & \textbf{Illusion} & 21.82 & \textbf{72.74} & 22.15 & 23.36 & 11.05 & \textbf{33.25} & 8.12 & 4.24 & 25.27 & 19.95 & 7.32 & 8.12 \\
\multicolumn{1}{|c|}{} & \textbf{Filtered} & 63.82 & \textbf{83.44} & 64.37 & 68.23 & 40.57 & 54.20 & 31.98 & 33.26 & 85.73 & 59.92 & 55.47 & 56.40 \\ \hline
\multicolumn{1}{|c|}{\multirow{3}{*}{\textbf{LLaVA}}} & \textbf{Raw} & 55.46 & 83.52 & 56.66 & 59.75 & 38.63 & 60.56 & 39.34 & 38.38 & \textbf{100} & \textbf{100} & \textbf{100} & \textbf{100} \\
\multicolumn{1}{|c|}{} & \textbf{Illusion} & 9.02 & 0.82 & 9.09 & 1.50 & 9.00 & 0.82 & 9.01 & 1.51 & 13.91 & 45.61 & 13.91 & 10.42 \\
\multicolumn{1}{|c|}{} & \textbf{Filtered} & 9.02 & 0.82 & 9.09 & 1.50 & 13.73 & 26.90 & 13.72 & 7.55 & 51.18 & 90.09 & 51.18 & 52.83 \\ \hline
\multicolumn{1}{|c|}{\multirow{3}{*}{\textbf{Kosmos-2}}} & \textbf{Raw} & 6.22 & 4.54 & 0.49 & 0.82 & 12.07 & 5.82 & 6.97 & 2.99 & 92.40 & 67.79 & 66.00 & 66.35 \\
\multicolumn{1}{|c|}{} & \textbf{Illusion} & 8.94 & 1.71 & 8.36 & 2.29 & 8.60 & 0.39 & 3.89 & 0.71 & 8.64 & 2.30 & 1.46 & 0.71 \\
\multicolumn{1}{|c|}{} & \textbf{Filtered} & 8.86 & 0.74 & 8.18 & 1.36 & 9.23 & 0.78 & 8.33 & 1.42 & 23.00 & 25.11 & 15.81 & 13.17 \\ \hline
\multicolumn{1}{|c|}{\multirow{3}{*}{\textbf{MiniGPT-V2}}} & \textbf{Raw} & 61.59 & 76.60 & 61.63 & 62.76 & 10.42 & 31.02 & 10.27 & 2.38 & \textbf{100} & \textbf{100} & \textbf{100} & \textbf{100} \\
\multicolumn{1}{|c|}{} & \textbf{Illusion} & 15.75 & 48.66 & 16.02 & 13.23 & 9.16 & 0.84 & 9.01 & 1.54 & 12.19 & \textbf{50.70} & 12.18 & 8.23 \\
\multicolumn{1}{|c|}{} & \textbf{Filtered} & 36.67 & 64.23 & 36.9 & 41.65 & 9.63 & 1.45 & 8.70 & 2.06 & 58.55 & 83.28 & 58.55 & 59.97 \\ \hline
\multicolumn{1}{|c|}{\multirow{3}{*}{\textbf{CLIP}}} & \textbf{Raw} & 25.97 & 28.01 & 27.70 & 20.14 & 41.15 & 47.13 & 41.33 & 37.54 & \textbf{100} & \textbf{100} & \textbf{100} & \textbf{100} \\
\multicolumn{1}{|c|}{} & \textbf{Illusion} & 15.26 & 18.07 & 15.32 & 13.02 & 13.89 & 27.99 & 13.92 & 9.20 & \textbf{42.64} & 46.20 & \textbf{42.64} & \textbf{39.56} \\
\multicolumn{1}{|c|}{} & \textbf{Filtered} & 21.16 & 21.17 & 21.63 & 18.93 & 44.75 & 50.50 & 44.74 & 41.28 & 85.45 & 86.37 & 85.45 & 85.19 \\ \hline
\multicolumn{1}{|c|}{\multirow{3}{*}{\textbf{BLIP}}} & \textbf{Raw} & 29.31 & 31.65 & 29.57 & 28.62 & 60.16 & 64.45 & 60.17 & 55.82 & \textbf{100} & \textbf{100} & \textbf{100} & \textbf{100} \\
\multicolumn{1}{|c|}{} & \textbf{Illusion} & 14.44 & 21.05 & 14.45 & 11.98 & 11.84 & 11.09 & 12.02 & 9.72 & 31.90 & 42.26 & 31.90 & 31.45 \\
\multicolumn{1}{|c|}{} & \textbf{Filtered} & 16.57 & 22.38 & 17.33 & 15.19 & 40.65 & 53.58 & 41.04 & 37.05 & 89.90 & 90.66 & 89.90 & 88.56 \\ \hline
\multicolumn{1}{|c|}{\multirow{3}{*}{\textbf{BLIP-2}}} & \textbf{Raw} & 61.68 & 70.26 & 62.85 & 61.81 & 64.41 & 70.10 & 64.40 & 63.84 & \textbf{100} & \textbf{100} & \textbf{100} & \textbf{100} \\
\multicolumn{1}{|c|}{} & \textbf{Illusion} & 15.67 & 15.83 & 15.65 & 12.85 & 11.69 & 22.68 & 11.47 & 9.49 & 32.64 & 41.46 & 32.64 & 30.05 \\
\multicolumn{1}{|c|}{} & \textbf{Filtered} & 40.20 & 43.30 & 40.84 & 37.77 & 45.15 & \textbf{62.46} & \textbf{45.19} & \textbf{45.08} & \textbf{93.73} & \textbf{94.54} & \textbf{93.73} & \textbf{94.23} \\ \hline
\multicolumn{1}{|c|}{\textbf{Human}} & \textbf{Illusion} & 96.69 & 97.22 & 96.05 & 96.43 & 74.6 & 72.81 & 73.63 & 72.85 & 93.03 & 92.86 & 90.88 & 91.5 \\ \hline
\end{tabular}
\begin{tablenotes}
  \item[*] The coverage of GPT-4o on different datasets is as follows: Coverage on IllusionMNIST for Raw, Illusion, and Filtered is 22.27\%, 99.84\%, and 96.8\% respectively. Coverage on IllusionFashionMNIST for Raw, Illusion, and Filtered is 70.75\%, 99.92\%, and 99.45\% respectively. Coverage on IllusionAnimals for Raw, Illusion, and Filtered is 99.90\%, 99.55\%, and 99.91\% respectively.
  \end{tablenotes}
\end{threeparttable}

}
\end{table*}

\begin{table*}[]
\caption{Zero-shot OCR capability of Gemini and GPT-4o on IllusionChar dataset}
\label{tab: zero-shot ocr}
\centering
\resizebox{0.5\linewidth}{!}{
\begin{tabular}{cc|c|c|c|}
\cline{3-5}
 &  & \textbf{Gemini} & \textbf{GPT-4o} & \textbf{Human} \\ \hline
\multicolumn{1}{|c|}{\multirow{2}{*}{\textbf{Raw}}} & \textbf{WER} & 53.73 & \textbf{22.34} & - \\
\multicolumn{1}{|c|}{} & \textbf{CER} & 56.73 & \textbf{7.01} & - \\ \hline
\multicolumn{1}{|c|}{\multirow{2}{*}{\textbf{Illusion}}} & \textbf{WER} & 90.48 & \textbf{90.26} & 31.94 \\
\multicolumn{1}{|c|}{} & \textbf{CER} & 175.98 & \textbf{169.46} & 13.32 \\ \hline
\multicolumn{1}{|c|}{\multirow{2}{*}{\textbf{Filtered}}} & \textbf{WER} & 82.73 & \textbf{76.4} & - \\
\multicolumn{1}{|c|}{} & \textbf{CER} & \textbf{99.93} & 122.18 & - \\ \hline
\end{tabular}
}
\end{table*}

\begin{table*}[]
\caption{Performance of different fine-tuned models on different datasets}
\label{fine-tuned}
\centering
\resizebox{\linewidth}{!}{
\begin{tabular}{cc|cccc|cccc|cccc|}
\cline{3-14}
 &  & \multicolumn{4}{c|}{\textbf{IllusionMNIST}} & \multicolumn{4}{c|}{\textbf{IllusionFashionMNIST}} & \multicolumn{4}{c|}{\textbf{IllusionAnimals}} \\ \cline{3-14} 
 &  & \textbf{Accuracy} & \textbf{Precision} & \textbf{Recall} & \textbf{F1} & \textbf{Accuracy} & \textbf{Precision} & \textbf{Recall} & \textbf{F1} & \textbf{Accuracy} & \textbf{Precision} & \textbf{Recall} & \textbf{F1} \\ \hline
\multicolumn{1}{|c|}{\multirow{3}{*}{\textbf{LLaVA}}} & \textbf{Raw} & 78.99 & 83.75 & 78.95 & 79.16 & 45.66 & 59.45 & 46.2 & 45.85 & \textbf{100} & \textbf{100} & \textbf{100} & \textbf{100} \\
\multicolumn{1}{|c|}{} & \textbf{Illusion} & 40.11 & 38.44 & 33.36 & 33.33 & 12.79 & 13.81 & 12.88 & 12.90 & 32.73 & 36.44 & 32.73 & 33.39 \\
\multicolumn{1}{|c|}{} & \textbf{Filtered} & 70.55 & 76.71 & 70.29 & 67.84 & 35.04 & 33.96 & 35.18 & 30.98 & 87.91 & 89.79 & 87.91 & 87.92 \\ \hline
\multicolumn{1}{|c|}{\multirow{3}{*}{\textbf{CLIP}}} & \textbf{Raw} & \textbf{92.25} & \textbf{93.24} & \textbf{92.22} & \textbf{92.24} & \textbf{84.20} & \textbf{85.85} & \textbf{84.99} & \textbf{84.35} & 99.60 & 99.61 & 99.60 & 99.60 \\
\multicolumn{1}{|c|}{} & \textbf{Illusion} & \textbf{91.8} & \textbf{92.47} & \textbf{91.72} & \textbf{91.69} & \textbf{83.90} & \textbf{85.17} & \textbf{84.64} & \textbf{84.15} & \textbf{94.36} & \textbf{94.87} & \textbf{94.36} & \textbf{94.37} \\
\multicolumn{1}{|c|}{} & \textbf{Filtered} & \textbf{91.55} & \textbf{92.32} & \textbf{91.47} & \textbf{91.47} & \textbf{82.00} & \textbf{83.48} & \textbf{82.92} & \textbf{81.82} & 88.73 & 90.80 & 88.73 & 88.55 \\ \hline
\multicolumn{1}{|c|}{\multirow{3}{*}{\textbf{BLIP}}} & \textbf{Raw} & 12.89 & 12.90 & 12.76 & 11.40 & 51.74 & 64.70 & 51.91 & 47.54 & 98.60 & 98.64 & 98.60 & 98.61 \\
\multicolumn{1}{|c|}{} & \textbf{Illusion} & 20.43 & 23.45 & 20.30 & 18.55 & 57.70 & 68.00 & 57.82 & 53.94 & 94.27 & 94.35 & 94.27 & 94.28 \\
\multicolumn{1}{|c|}{} & \textbf{Filtered} & 19.77 & 17.78 & 19.52 & 17.33 & 58.64 & 67.75 & 58.85 & 54.79 & \textbf{94.36} & \textbf{94.63} & \textbf{94.36} & \textbf{94.35} \\ \hline
\multicolumn{1}{|c|}{\multirow{3}{*}{\textbf{BLIP-2}}} & \textbf{Raw} & 89.36 & 90.06 & 89.60 & 89.11 & 43.92 & 38.42 & 44.25 & 36.00 & 69.10 & 67.63 & 69.10 & 62.80 \\
\multicolumn{1}{|c|}{} & \textbf{Illusion} & 55.54 & 57.64 & 55.56 & 55.22 & 41.99 & 34.68 & 42.55 & 36.72 & 67.55 & 68.35 & 67.55 & 67.29 \\
\multicolumn{1}{|c|}{} & \textbf{Filtered} & 86.79 & 87.70 & 86.91 & 86.57 & 46.88 & 43.96 & 47.43 & 41.96 & 89.45 & 90.30 & 89.45 & 89.33 \\ \hline
\end{tabular}
}

\end{table*}

\section{Results}
In our study, we observe the performance of various models on different datasets, both with and without the application of illusions and filters. Let's break down the findings for each dataset:
\subsection{IllusionMNIST Dataset}
\begin{itemize}
    \item As illustrated in Table \ref{tab: zero-shot}, when applying illusions to the original images, we notice a significant drop in the performance of the models. This indicates that VLMs struggle to interpret illusions in images.
    \item Among all the models, GPT-4o achieves the best results on the IllusionMNIST dataset, surpassing other models in both the illusion and filter-applied scenarios.
    \item Additionally, when we apply our simple filter to the images, we observe that almost all models, except for Kosmos-2 and LLaVA, show improved performance compared to the scenario without the filter. This demonstrates the effectiveness of our approach to this dataset.
    \item After fine-tuning the models on the provided training dataset, we observe a noticeable improvement in their performance, as demonstrated in Table \ref{fine-tuned}.
\end{itemize}

\subsection{IllusionFashionMNIST Dataset}
\begin{itemize}
    \item Also, in this case, we see that applying illusion causes a drop in results.
    \item Prior to applying filters, GPT-4o exhibits the best results on the illusory images of the IllusionFashionMNIST dataset.
    \item After applying the filter, BLIP-2 achieves the best performance.
    \item Similarly to the previous dataset, we observe that all models show improved performance when the filter is applied to the images.  This suggests that the filter has a positive impact on the models' ability to interpret the illusions.
    \item Likewise, we observed an improvement in performance for all models after fine-tuning.
\end{itemize}

\subsection{IllusionAnimals Dataset}
\begin{itemize}
    \item Similarly, in this case, we observe that applying illusion causes a drop in results.
    \item Before applying filters, CLIP displays the highest performance on the illusory images of the IllusionAnimals dataset.
    \item After applying the filter, BLIP-2 yields the best results, even better than humans.
    \item Once again, the application of the filter results in an increase in the performance of all models on the illusory images. This further emphasizes the effectiveness of our solution in improving model performance on illusory images.
    \item Similarly, we observed an enhancement in performance for all models after the process of fine-tuning.
\end{itemize}

\subsection{IllusionChar Dataset}
\begin{itemize}
    \item Similarly, in this case, the application of the illusion technique leads to a drop in results, as demonstrated in Table \ref{tab: zero-shot ocr}.
    \item Before applying filters, GPT-4o demonstrates better performance than Gemini.
    \item After applying the filter, performance improves for both models, highlighting the effectiveness of our solution. GPT-4o exhibits enhanced performance in terms of WER, while Gemini showcases superior performance in CER.
\end{itemize}
Overall, our study highlights that the interpretation of illusions in images poses a challenge for VLMs. However, the application of filters can significantly improve model performance. You can see better visualization of results in Appendix \ref{appendix:results_visualization}. One interesting observation is that, although we fine-tuned our models on illusory data, the performance of our model on raw images increased in some cases.

\section{Conclusion and Future Work}
In conclusion, despite the significant advancements in multimodal models, there remain notable disparities in how these models perceive and interpret images compared to human perception. One specific area where such disparities exist is in illusory images. To address this, we introduce IllusionMNIST, IllusionFashionMNIST, IllusionAnimals, and IllusionChar datasets, which serve as valuable benchmarks for evaluating the performance of VLMs on illusory images. Additionally, we propose a straightforward yet effective approach to enhance the performance of these models on such images. This involves applying a Gaussian and blur low-pass filter and converting the images to grayscale. Implementing this method demonstrates notable performance improvements. Furthermore, we suggest the possibility of fine-tuning VLMs specifically for illusory images.\\
Looking ahead, there is ample room for future exploration and development. One potential avenue is the utilization of adaptive and learnable filters tailored specifically for these types of images. Furthermore, expanding the collection of more comprehensive and diverse datasets for illusory images would be beneficial, while our current study has primarily focused on image classification and OCR capability. It would be worthwhile to investigate the impact of in-context learning on these types of images. By delving into these areas, we can further bridge the gap between multimodal models and human perception, ultimately advancing the field of image interpretation and analysis. Novel architectures specifically designed to address the challenges posed by illusory images can be explored. The application of adversarial training techniques to improve the models' robustness against illusory images can be investigated. Human feedback and input can be incorporated into the training process to refine the models' understanding of illusory images, bridging the gap between model and human perception. Quantitative evaluation metrics specifically tailored for assessing performance on illusory images can be developed.

\section{Limitations}
Our work primarily focuses on the classification and optical character recognition (OCR) aspects of illusory images, while our goal was to develop a comprehensive dataset for illusory images to support general Visual Question Answering (VQA) task. However, we encountered several limitations:
\begin{itemize}
    \item \textbf{Limited Scope of Illusory Images} Our work primarily focuses on illusions in images related to classification and OCR. While we aimed to create a comprehensive dataset for illusory images for general VQA tasks, there are certain limitations that need to be acknowledged.

    \item \textbf{Inability to Assess Color Changes} One limitation is that we were unable to assess the color of the images using the Visual Language Model (VLM) due to the nature of illusions. Applying illusions to images can alter the appearance of objects, making it challenging to accurately determine their original colors.

    \item \textbf{Single Large Objects} Another limitation is that our work mainly focuses on images containing just one large-sized object. This choice was made to simplify the dataset construction process and reduce the burden of evaluation. However, it is important to note that illusions can occur with multi-object scenes and objects of varying sizes. Exploring these scenarios could provide a more comprehensive understanding of illusions in images.

    \item \textbf{Reduced Task Difficulty} Constructing datasets in the classification and OCR formats helps streamline the evaluation process. However, this approach may inadvertently reduce the difficulty of the task. It allows models to potentially solve the classification task by randomly choosing one of the classes, rather than genuinely understanding the underlying illusions present in the images.
\end{itemize}
When considering the limitations of our study, it is important to acknowledge the hardware constraints that have influenced our approach. Specifically, we have chosen not to fine-tune our models on IllusionChar due to these limitations. By refraining from this step, we aim to explore the potential outcomes that may arise. By acknowledging these limitations, we can better understand the context and scope of our work and identify areas for future research and improvement.

\bibliographystyle{plainnat}  
\bibliography{main}


\appendix

\section{Task Examples}
\label{appendix: Task Examples}
\begin{figure*}[h]
    \centering
    \includegraphics[width=\textwidth]{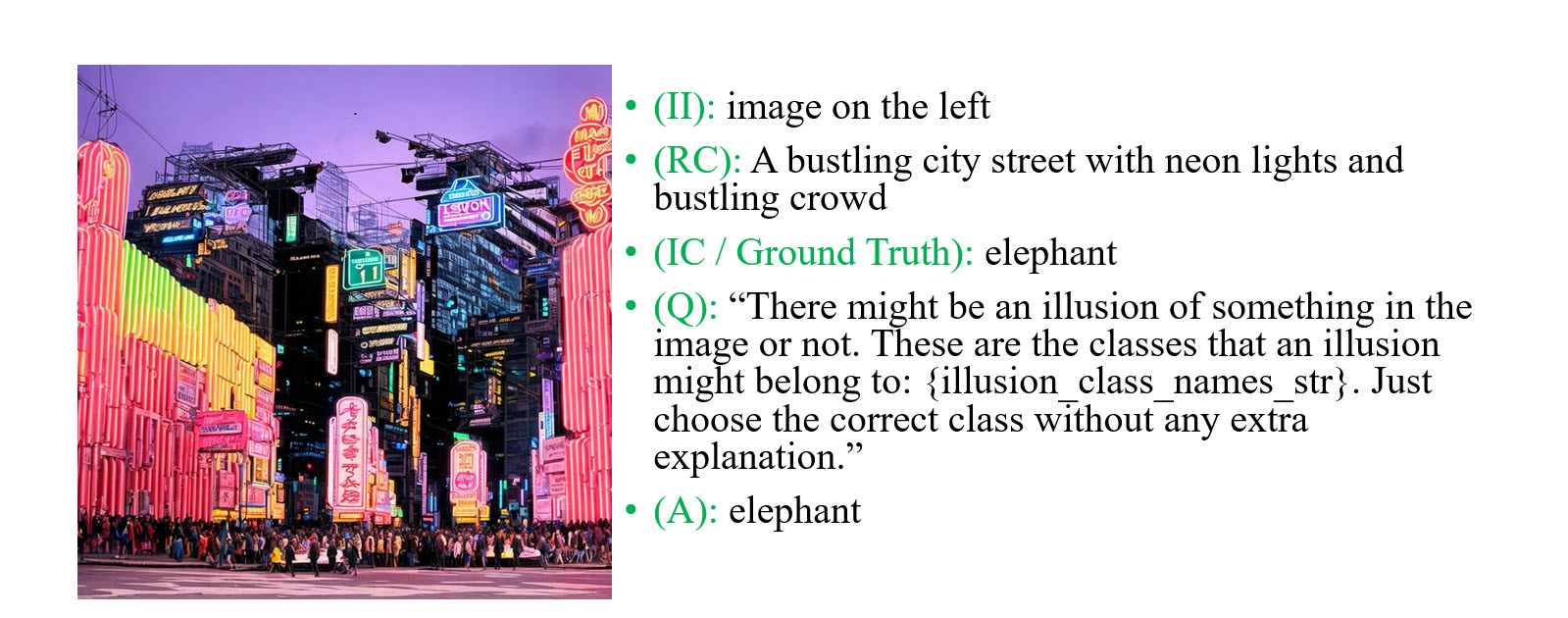}
    \caption{An example of task definition}
    \label{appendix: elephant}
\end{figure*}

\begin{figure*}[h]
    \centering
    \includegraphics[width=\textwidth]{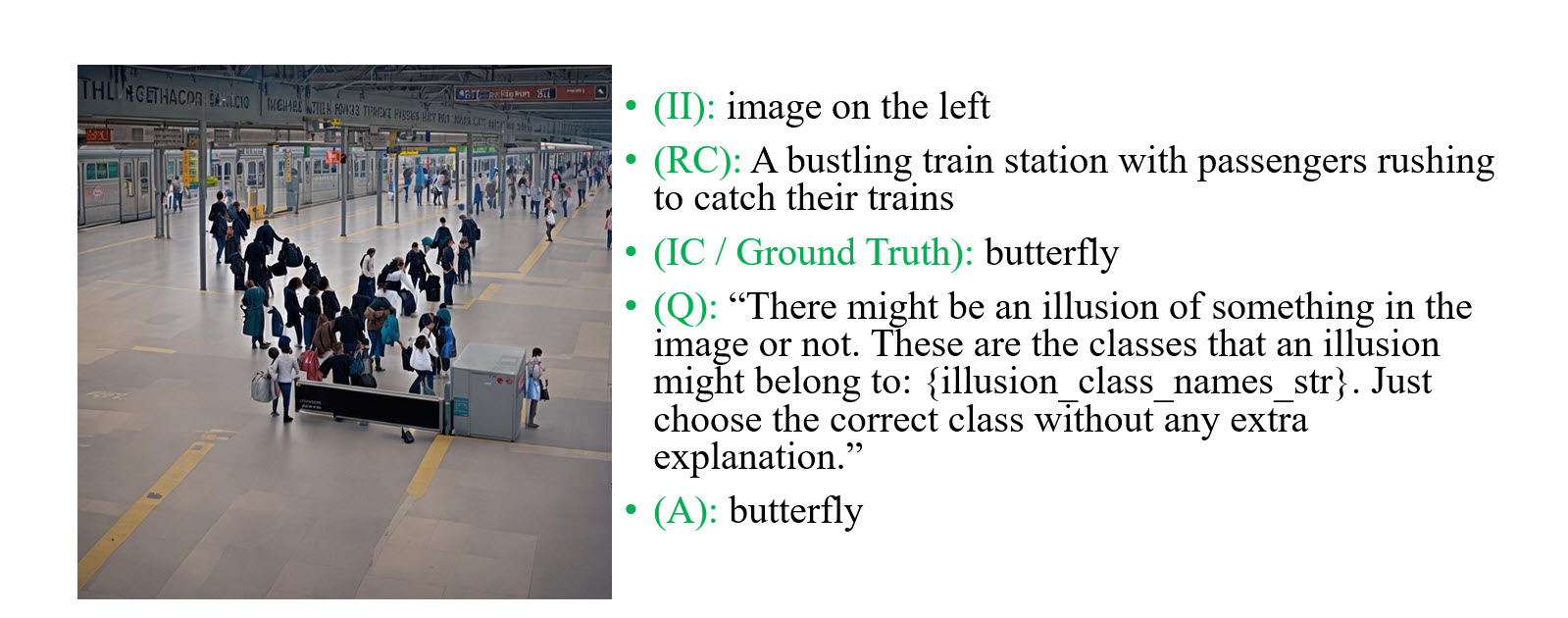}
    \caption{An example of task definition}
    \label{appendix: butterfly}
\end{figure*}

In Figures \ref{appendix: elephant}, \ref{appendix: butterfly}, you find some examples of our task.

\section{Dataset Examples}
\begin{figure*}[h]
    \centering
    \includegraphics[width=\textwidth]{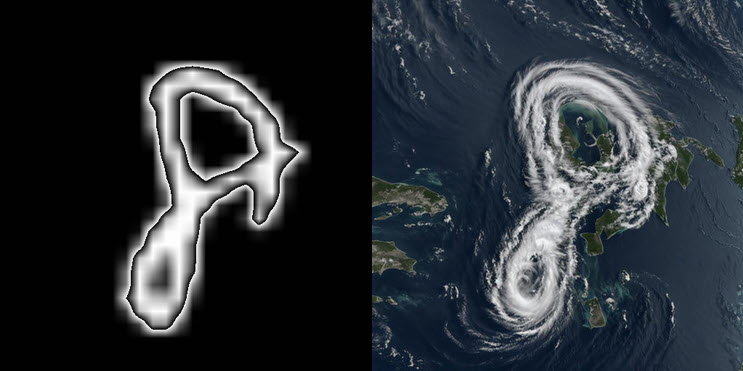}
    \caption{An example of IllusionMNIST}
    \label{appendix: illusionmnist_example}
\end{figure*}

\begin{figure*}[h]
    \centering
    \includegraphics[width=\textwidth]{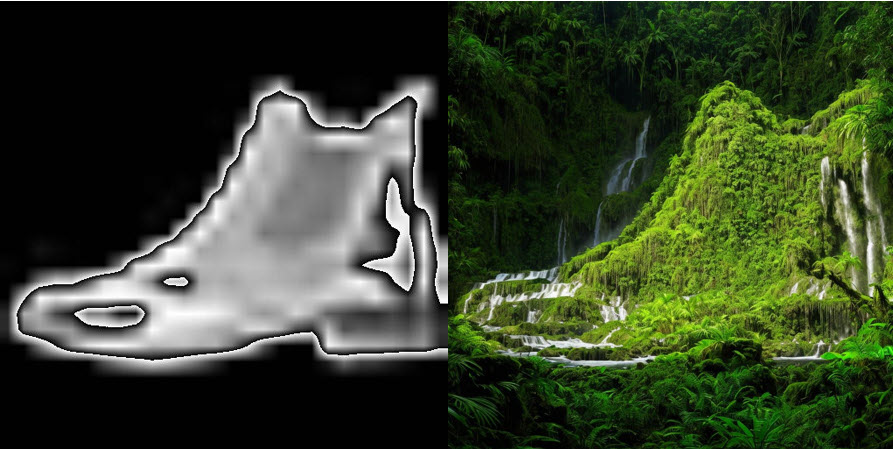}
    \caption{An example of IllusionFashionMNIST}
    \label{appendix: illusionfashionmnist_example}
\end{figure*}

\begin{figure*}[h]
    \centering
    \includegraphics[width=\textwidth]{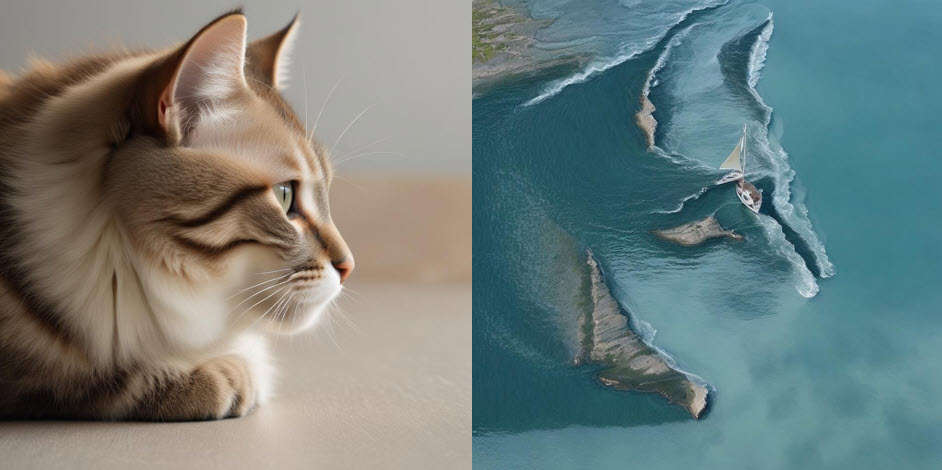}
    \caption{An example of IllusionAnimals}
    \label{appendix: illusionanimals_example}
\end{figure*}

\begin{figure*}[h]
    \centering
    \includegraphics[width=\textwidth]{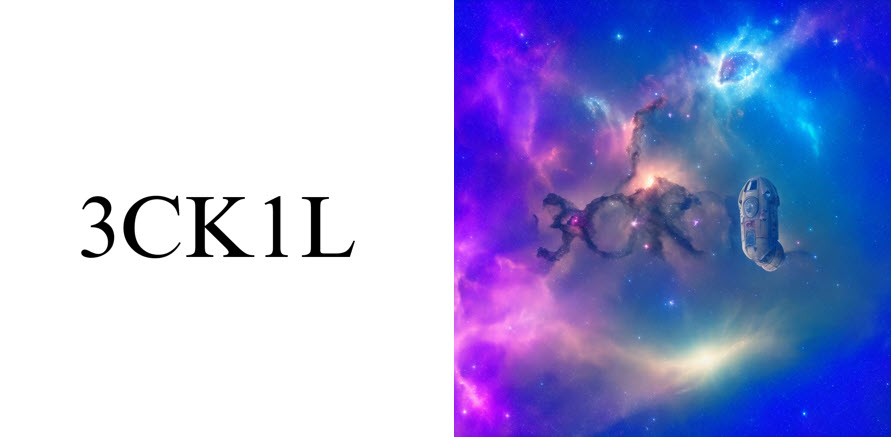}
    \caption{An example of IllusionChar}
    \label{appendix: illusionchar_example}
\end{figure*}

In Figures \ref{appendix: illusionmnist_example}, \ref{appendix: illusionfashionmnist_example}, \ref{appendix: illusionanimals_example}, and \ref{appendix: illusionchar_example}, you can find examples from each dataset.

\section{Human Evaluation}
\label{appendix:human}
For the human evaluation study, we present annotators with illusory images along with choices for three datasets: IllusionFashionMNIST, IllusionMNIST, and IllusionAnimals. Annotators are first asked to choose the label they suppose to be true from a set of given labels. After selecting the label, they are then shown the true label and asked to determine whether the true label is present in the image by answering with "yes" or "no." For the IllusionChar dataset, we provide images and ask annotators to type the character sequence they see. To ensure the validity of our dataset, we presented participants with the true sequence of characters in the image and asked them to confirm if they could perceive it. The results of this human evaluation are presented in Tables \ref{tab:human_val_1} and \ref{tab:human_val_2}.

\begin{table*}[h]
\caption{Human performance on 3 datasets: IllusionMNIST, IllusionFashionMNIST and IllusionAnimals}
\label{tab:human_val_1}
\centering
\resizebox{\textwidth}{!}{%
\begin{tabular}{cc|c|c|c|c|}
\cline{3-6}
\multicolumn{2}{c|}{} & \textbf{Annotator 1} & \textbf{Annotator 2} & \textbf{Annotator 3} & \textbf{Annotator 4} \\ \hline
\multicolumn{2}{|c|}{\textbf{Age}} & 24 & 23 & 22 & 22 \\ \hline
\multicolumn{2}{|c|}{\textbf{Gender}} & M & F & M & M \\ \hline
\multicolumn{1}{|c|}{\multirow{4}{*}{\textbf{IllusionMNIST}}} & \textbf{Accuracy} & 98.35 & 95.04 & - & 96.69 \\
\multicolumn{1}{|c|}{} & \textbf{Precision} & 98.74 & 95.86 & - & 97.39 \\
\multicolumn{1}{|c|}{} & \textbf{Recall} & 97.73 & 94.41 & - & 96.02 \\
\multicolumn{1}{|c|}{} & \textbf{F1} & 98.05 & 94.77 & - & 96.47 \\ \hline
\multicolumn{1}{|c|}{\multirow{4}{*}{\textbf{IllusionFashionMNIST}}} & \textbf{Accuracy} & 73.02 & 69.84 & 80.95 & - \\
\multicolumn{1}{|c|}{} & \textbf{Precision} & 71.65 & 66.69 & 80.02 & - \\
\multicolumn{1}{|c|}{} & \textbf{Recall} & 71.55 & 69.27 & 80.07 & - \\
\multicolumn{1}{|c|}{} & \textbf{F1} & 70.95 & 67.33 & 79.34 & - \\ \hline
\multicolumn{1}{|c|}{\multirow{4}{*}{\textbf{IllusionAnimals}}} & \textbf{Accuracy} & 93.64 & - & 94.55 & 90.91 \\
\multicolumn{1}{|c|}{} & \textbf{Precision} & 92.95 & - & 95.38 & 92.98 \\
\multicolumn{1}{|c|}{} & \textbf{Recall} & 91.99 & - & 91.70 & 88.96 \\
\multicolumn{1}{|c|}{} & \textbf{F1} & 92.18 & - & 92.14 & 89.99 \\ \hline
\end{tabular}%
}

\end{table*}

\begin{table*}[h]
\caption{Human performance on IllusionChar dataset}
\label{tab:human_val_2}
\centering
\begin{tabular}{c|c|c|cc|}
\cline{2-5}
\multirow{2}{*}{} & \multirow{2}{*}{\textbf{Age}} & \multirow{2}{*}{\textbf{Gender}} & \multicolumn{2}{c|}{\textbf{IllusionChar}} \\ \cline{4-5} 
 &  &  & \textbf{WER} & \textbf{CER} \\ \hline
\multicolumn{1}{|c|}{\textbf{Annotator 1}} & 24 & M & - & - \\ \hline
\multicolumn{1}{|c|}{\textbf{Annotator 2}} & 23 & F & 32.22 & 14.11 \\ \hline
\multicolumn{1}{|c|}{\textbf{Annotator 3}} & 22 & M & 32.50 & 12.07 \\ \hline
\multicolumn{1}{|c|}{\textbf{Annotator 4}} & 22 & M & 31.11 & 13.78 \\ \hline
\end{tabular}%

\end{table*}

\section{Datasets' Pie Charts}
\label{appendix:pie_chart}
\begin{figure}[h!]
    \centering
    \begin{minipage}[b]{0.45\linewidth}
        \includegraphics[width=\linewidth]{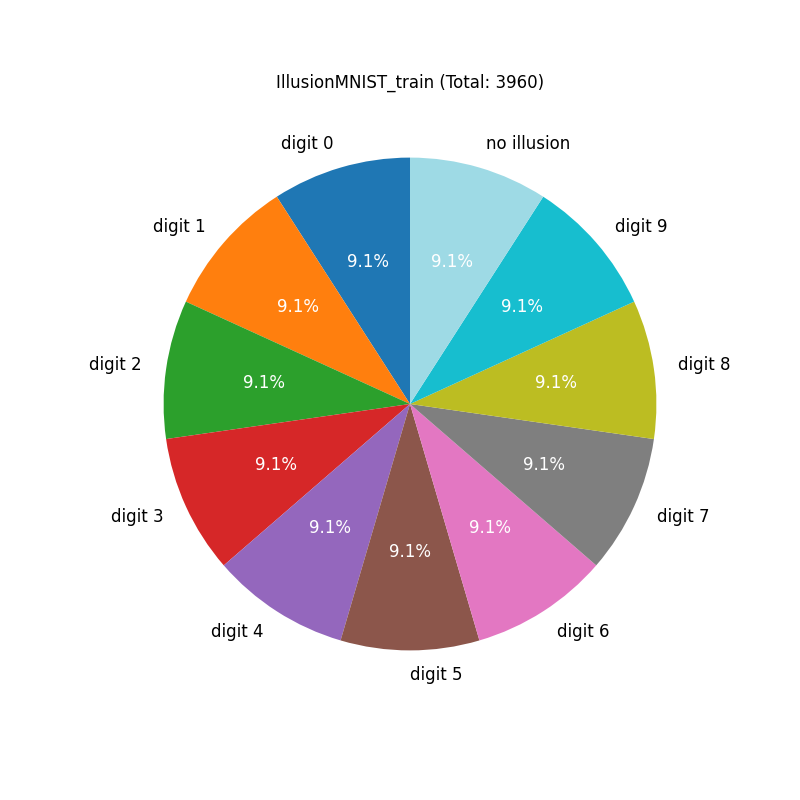}
        \caption{IllusionMNIST train split}
        \label{fig:illusionMNIST_train_pie}
    \end{minipage}
    \hspace{0.05\linewidth}
    \begin{minipage}[b]{0.45\linewidth}
        \includegraphics[width=\linewidth]{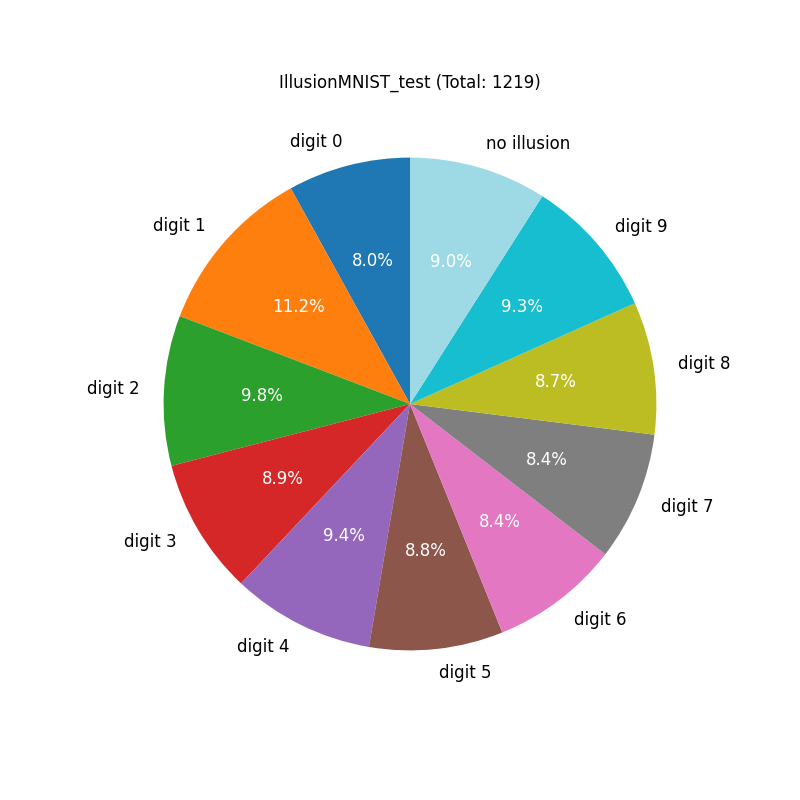}
        \caption{IllusionMNIST test split}
        \label{fig:illusionMNIST_test_pie}
    \end{minipage}
\end{figure}

\begin{figure}[h!]
    \centering
    \begin{minipage}[b]{0.45\linewidth}
        \includegraphics[width=\linewidth]{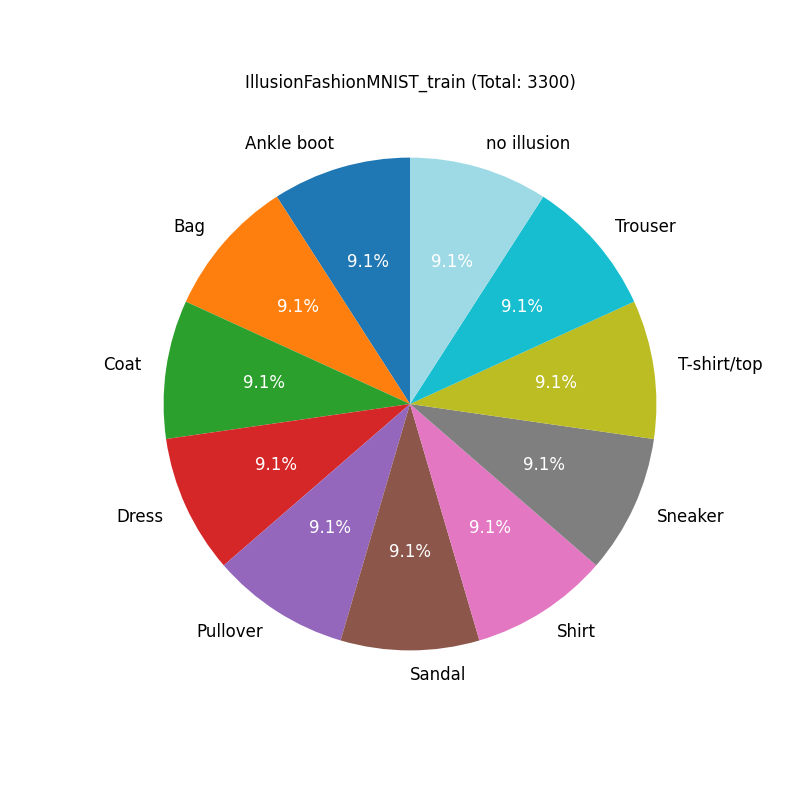}
        \caption{IllusionFashionMNIST train split}
        \label{fig:illusionFashionMNIST_train_pie}
    \end{minipage}
    \hspace{0.05\linewidth}
    \begin{minipage}[b]{0.45\linewidth}
        \includegraphics[width=\linewidth]{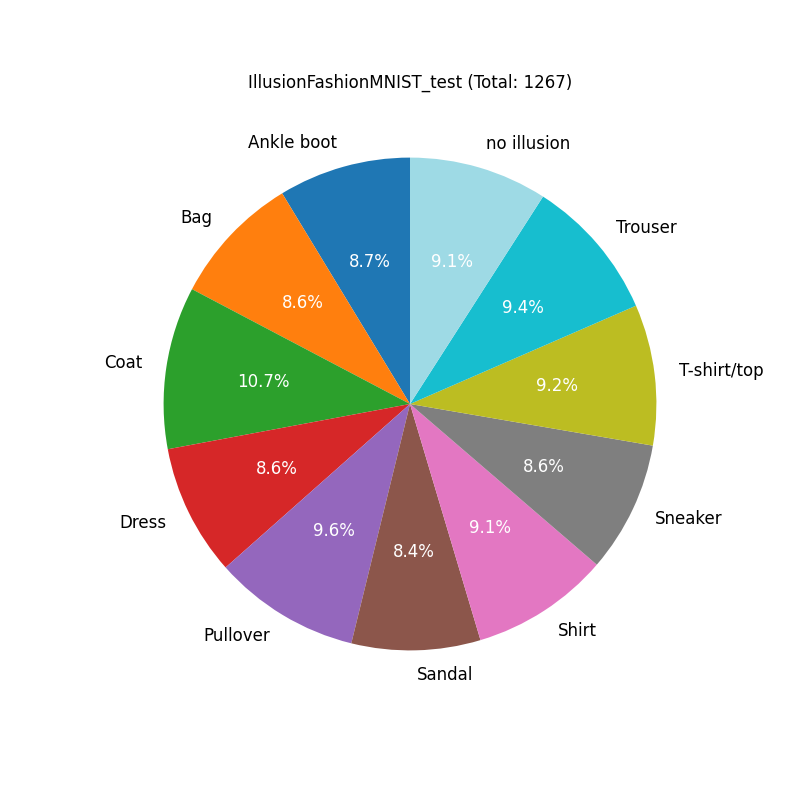}
        \caption{IllusionFashionMNIST test split}
        \label{fig:illusionFashionMNIST_test_pie}
    \end{minipage}
\end{figure}

\begin{figure}[h!]
    \centering
    \begin{minipage}[b]{0.45\linewidth}
        \includegraphics[width=\linewidth]{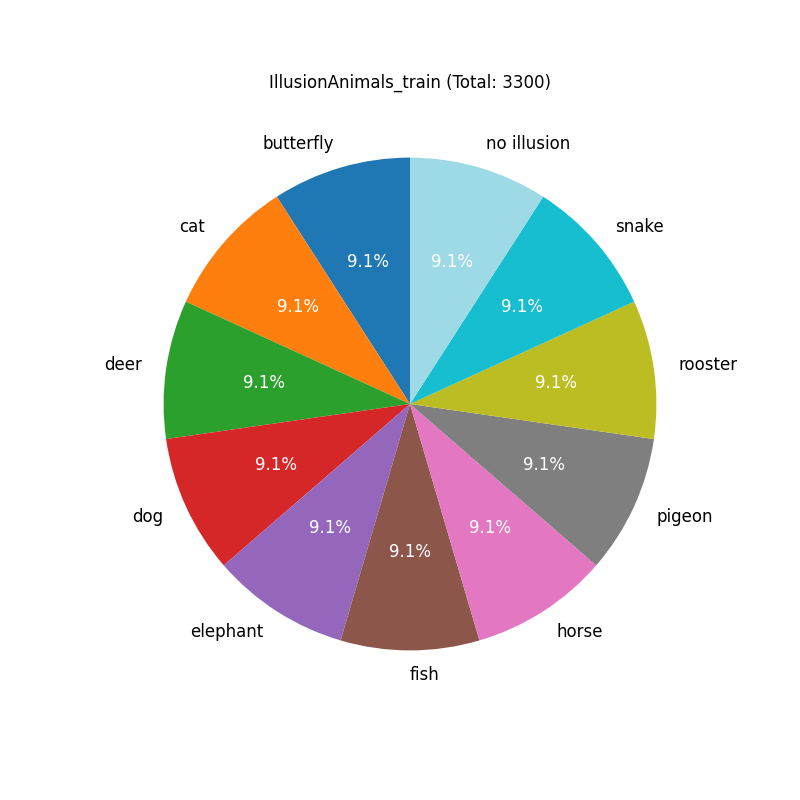}
        \caption{IllusionAnimals train split}
        \label{fig:illusionAnimals_train_pie}
    \end{minipage}
    \hspace{0.05\linewidth}
    \begin{minipage}[b]{0.45\linewidth}
        \includegraphics[width=\linewidth]{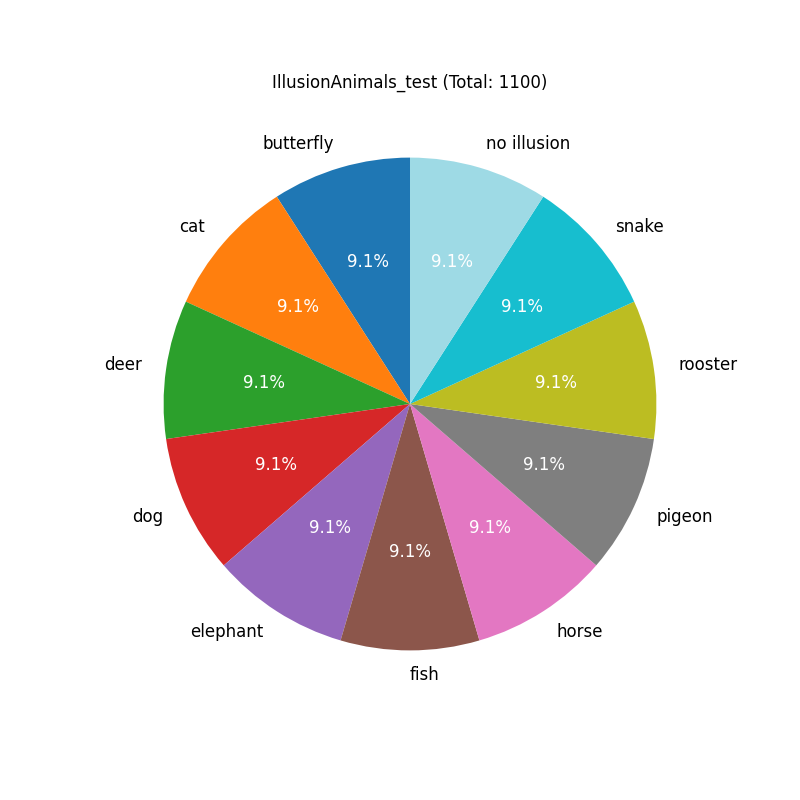}
        \caption{IllusionAnimals test split}
        \label{fig:illusionAnimals_test_pie}
    \end{minipage}
\end{figure}

The pie charts offer an overview of the proportion of each class present in the datasets. The datasets included are IllusionMNIST, IllusionFashionMNIST, and IllusionAnimals, each analyzed separately for their training and testing splits. For the IllusionMNIST dataset, see the train split in Figure \ref{fig:illusionMNIST_train_pie} and the test split in Figure \ref{fig:illusionMNIST_test_pie}. The IllusionFashionMNIST dataset is shown in Figure \ref{fig:illusionFashionMNIST_train_pie} for the train split and Figure \ref{fig:illusionFashionMNIST_test_pie} for the test split. Finally, the IllusionAnimals dataset splits are depicted in Figure \ref{fig:illusionAnimals_train_pie} for the train split and Figure \ref{fig:illusionAnimals_test_pie} for the test split.

\section{Results Visualization}
\label{appendix:results_visualization}
\begin{figure}[h!]
    \centering
    \includegraphics[width=0.9\columnwidth]{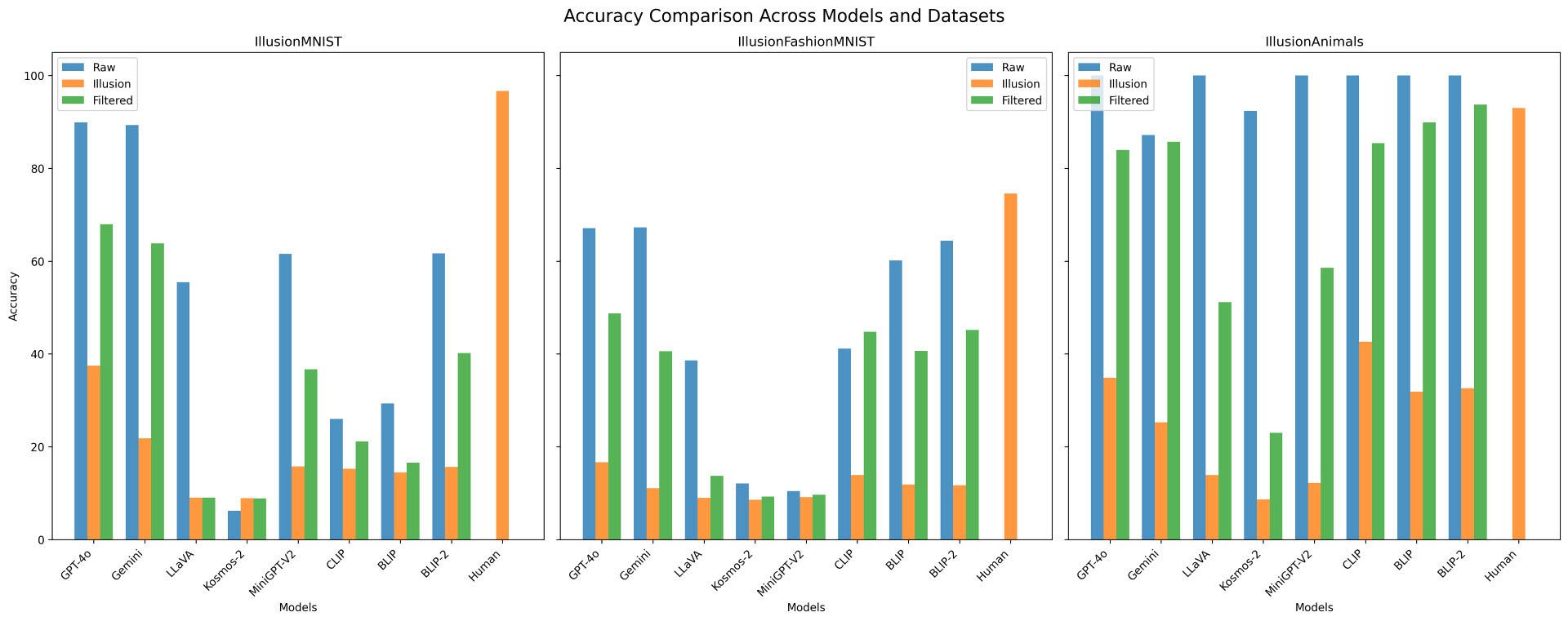}
    \caption{Visualization of zero-shot classification accuracies across various datasets}
    \label{fig:z_accuracy_comparison}
\end{figure}

\begin{figure}[h!]
    \centering
    \includegraphics[width=0.9\columnwidth]{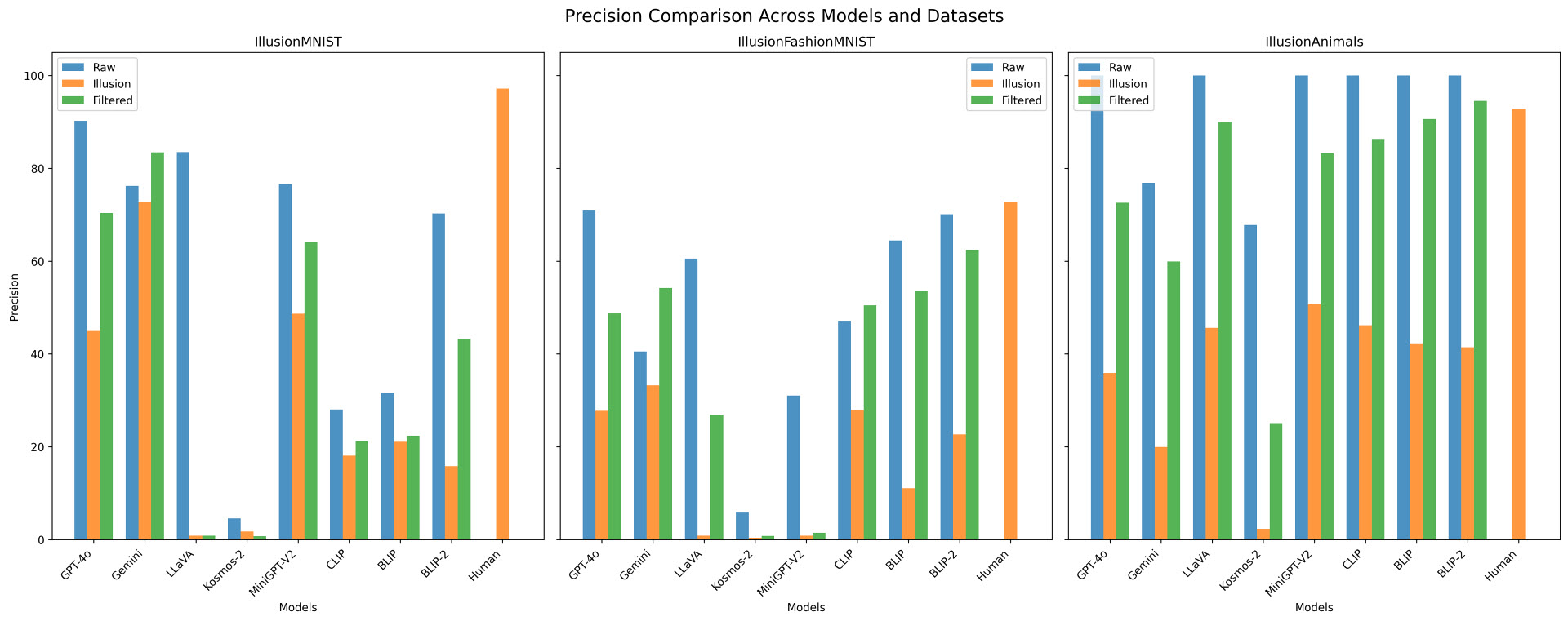}
    \caption{Visualization of zero-shot classification precisions across various datasets}
    \label{fig:z_precision_comparison}
\end{figure}

\begin{figure}[h!]
    \centering
    \includegraphics[width=0.9\columnwidth]{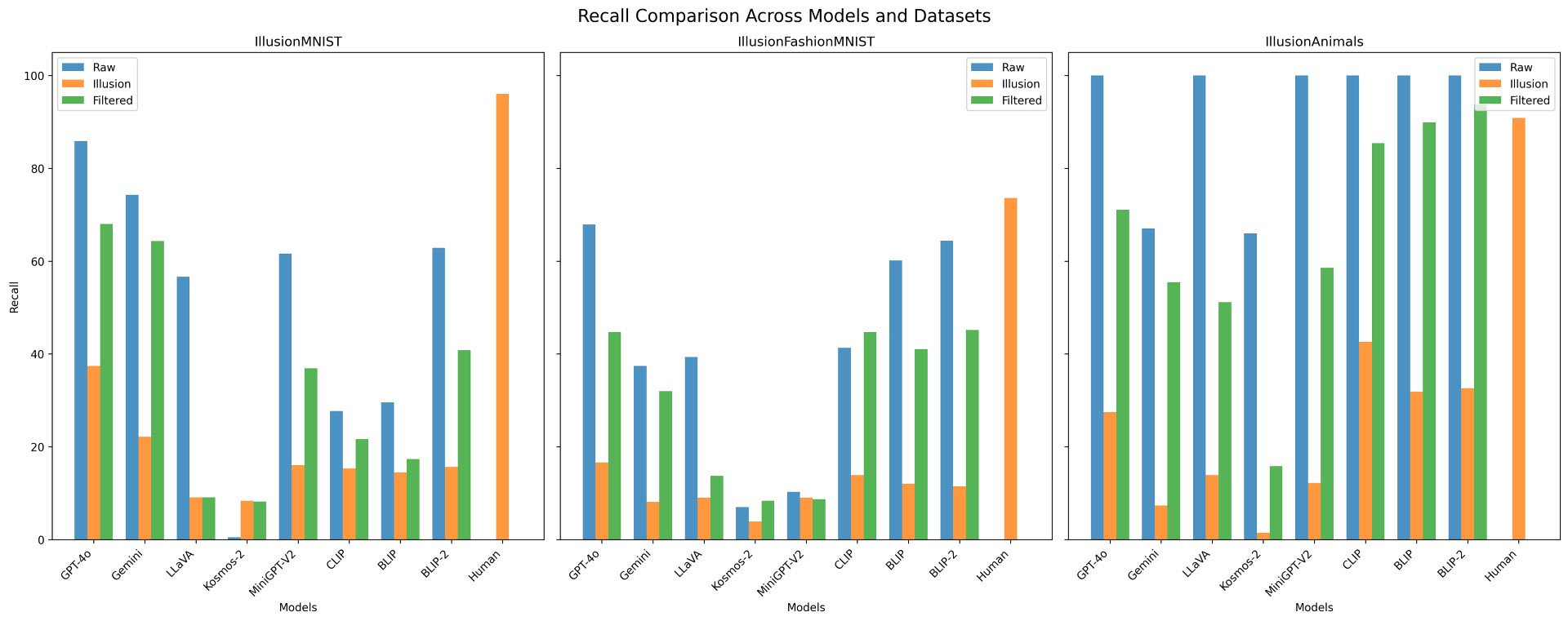}
    \caption{Visualization of zero-shot classification recalls across various datasets}
    \label{fig:z_recall_comparison}
\end{figure}

\begin{figure}[h!]
    \centering
    \includegraphics[width=0.9\columnwidth]{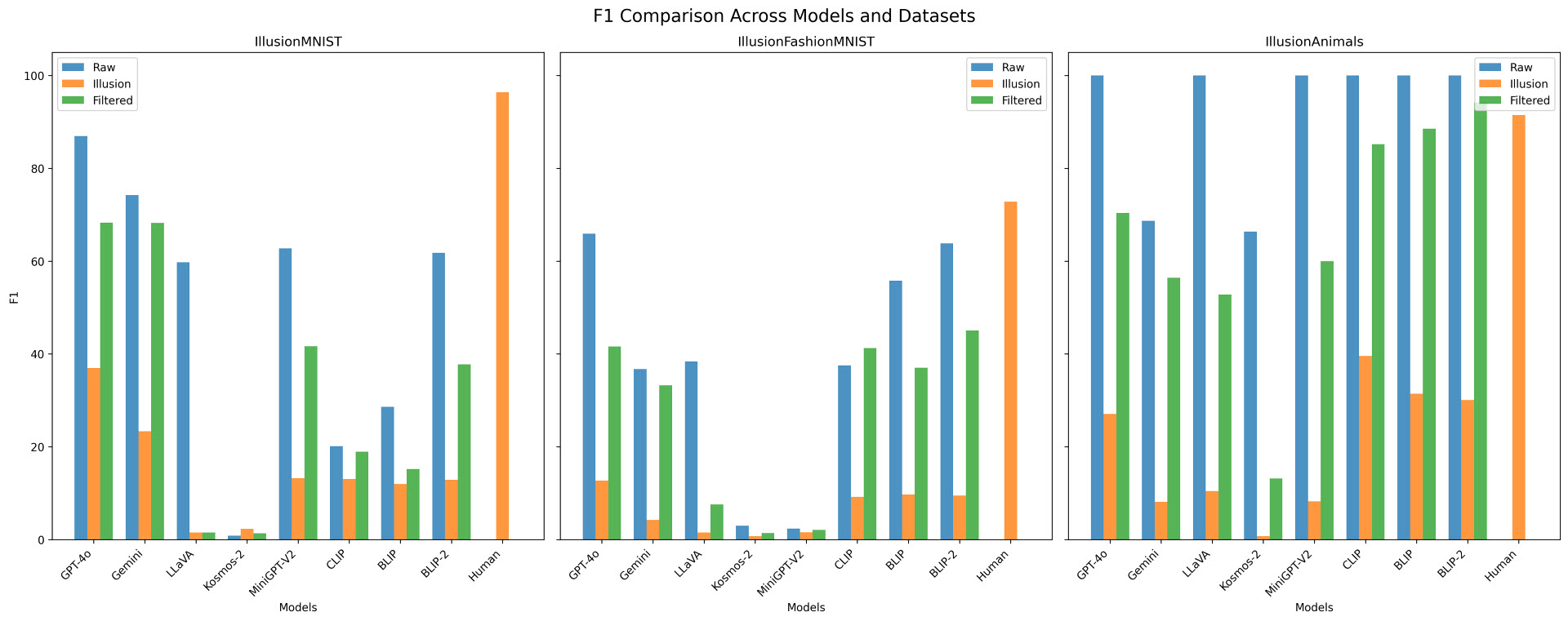}
    \caption{Visualization of zero-shot classification f1 scores across various datasets}
    \label{fig:z_f1_comparison}
\end{figure}

\begin{figure}[h!]
    \centering
    \includegraphics[width=0.9\columnwidth]{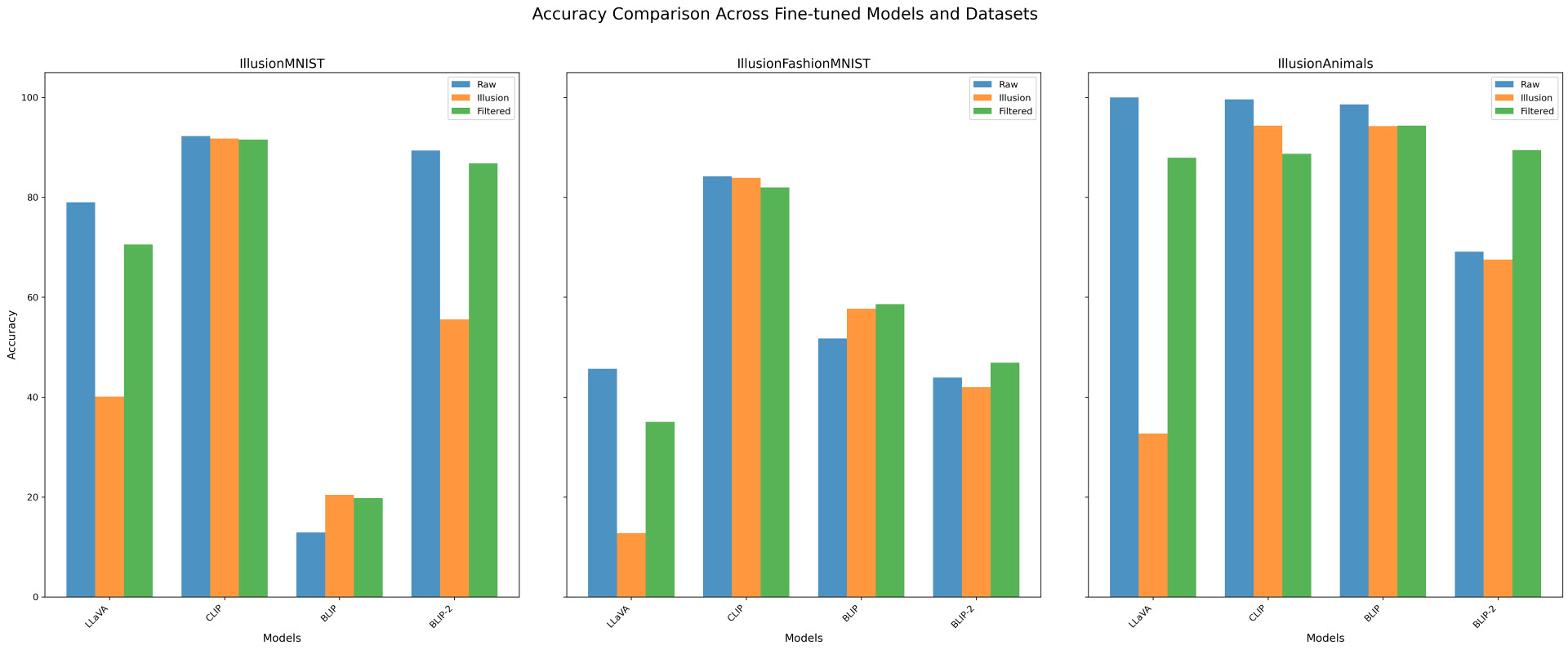}
    \caption{Visualization of fine-tuned classification accuracies across various datasets}
    \label{fig:accuracy_comparison}
\end{figure}

\begin{figure}[h!]
    \centering
    \includegraphics[width=0.9\columnwidth]{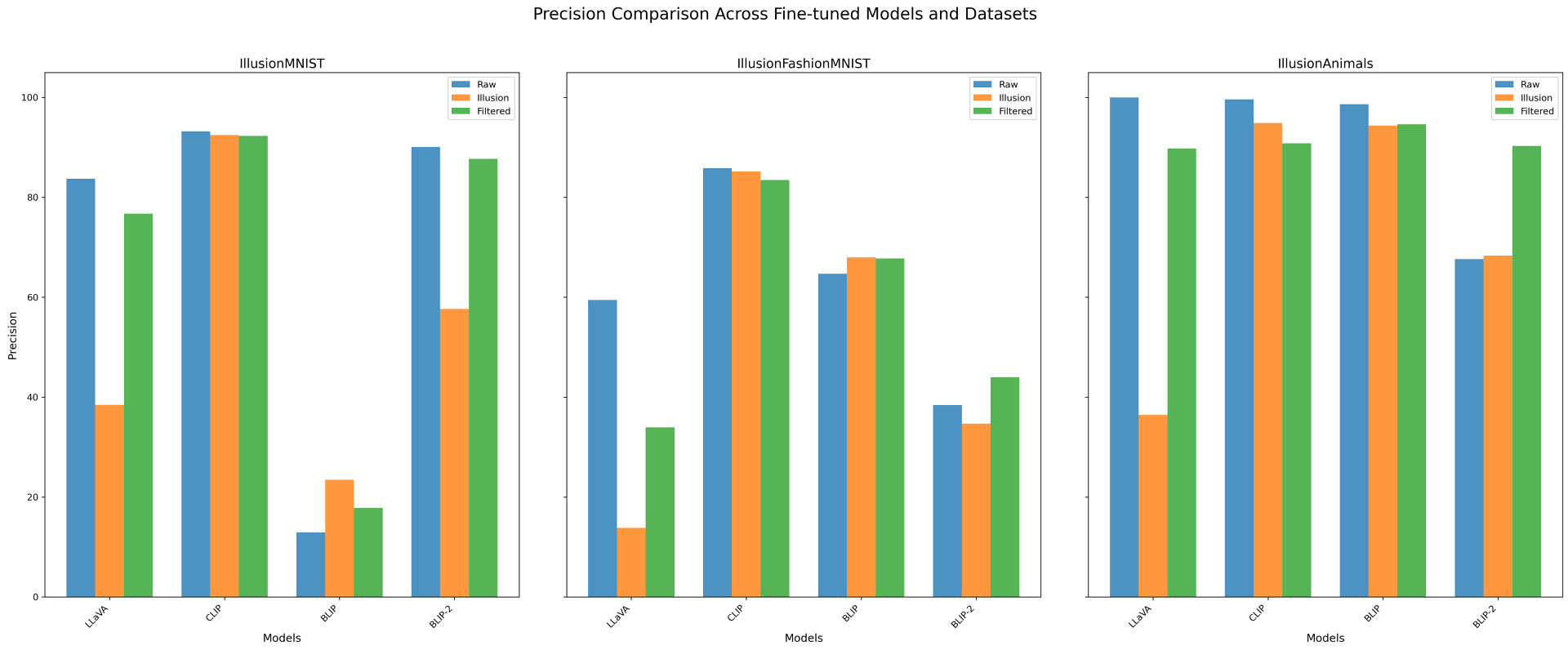}
    \caption{Visualization of fine-tuned classification precisions across various datasets}
    \label{fig:precision_comparison}
\end{figure}

\begin{figure}[h!]
    \centering
    \includegraphics[width=0.9\columnwidth]{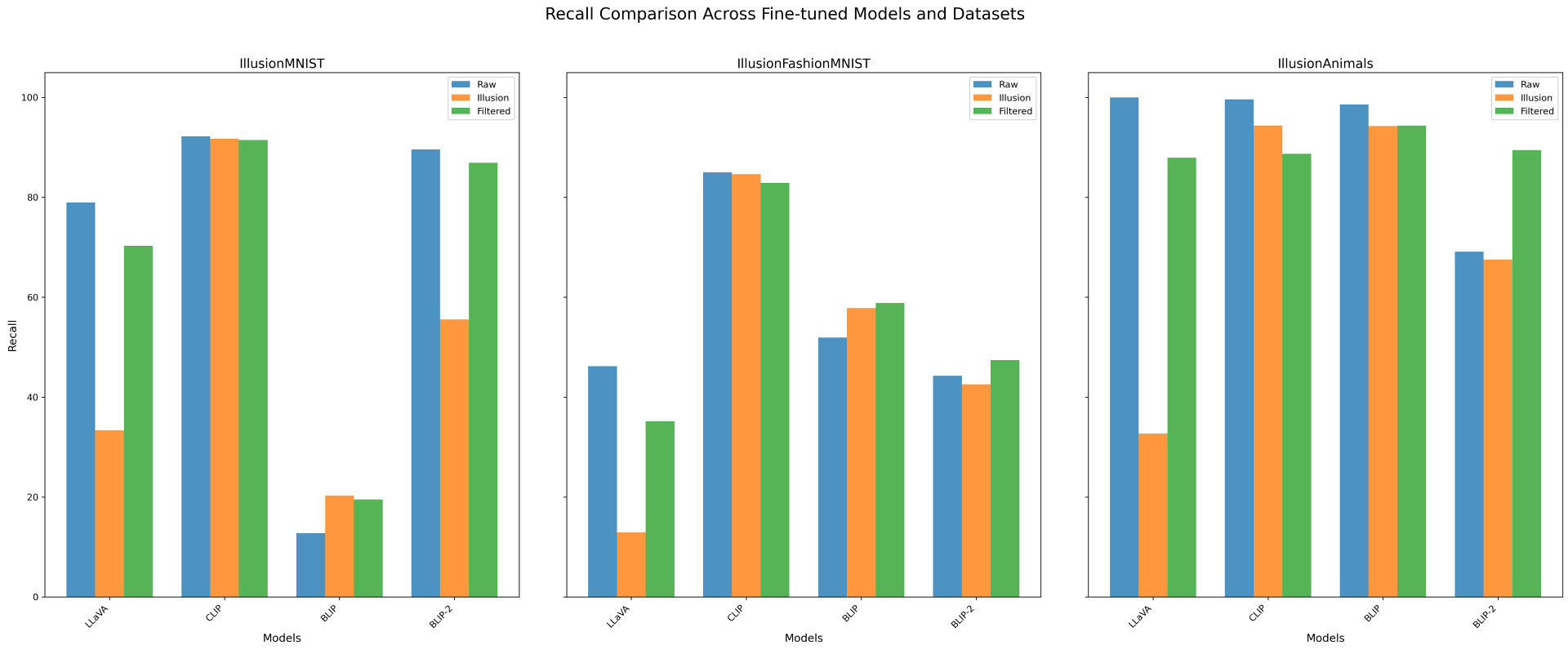}
    \caption{Visualization of fine-tuned classification recalls across various datasets}
    \label{fig:recall_comparison}
\end{figure}

\begin{figure}[h!]
    \centering
    \includegraphics[width=0.9\columnwidth]{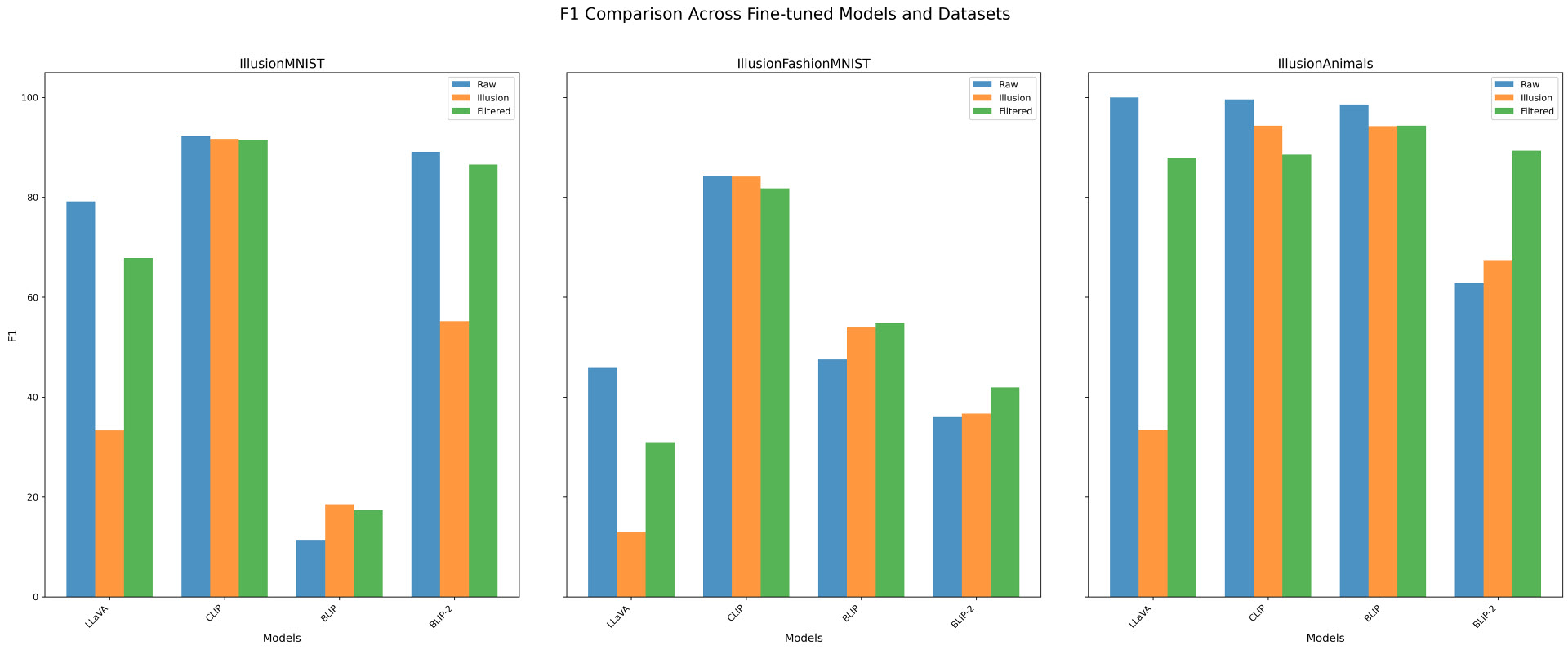}
    \caption{Visualization of fine-tuned classification f1 scores across various datasets}
    \label{fig:f1_comparison}
\end{figure}

\begin{figure}[h!]
    \centering
    \includegraphics[width=0.9\columnwidth]{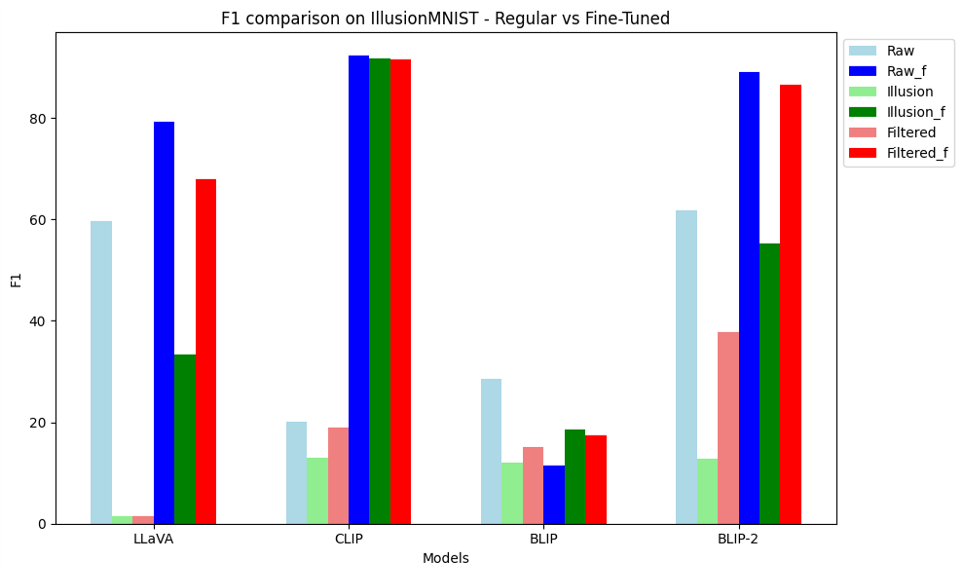}
    \caption{Comparison of zero-shot vs. fine-tuned classification f1 scores on IllusionMNIST dataset}
    \label{fig:f1_illusionmnist}
\end{figure}

\begin{figure}[h!]
    \centering
    \includegraphics[width=0.9\columnwidth]{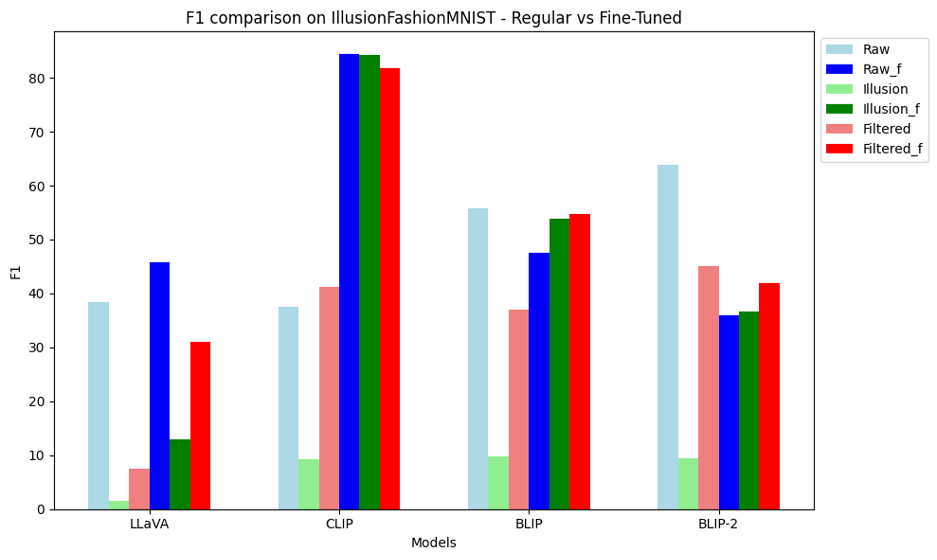}
    \caption{Comparison of zero-shot vs. fine-tuned classification f1 scores on IllusionFashionMNIST dataset}
    \label{fig:f1_illusionfashionmnist}
\end{figure}

\begin{figure}[h!]
    \centering
    \includegraphics[width=0.9\columnwidth]{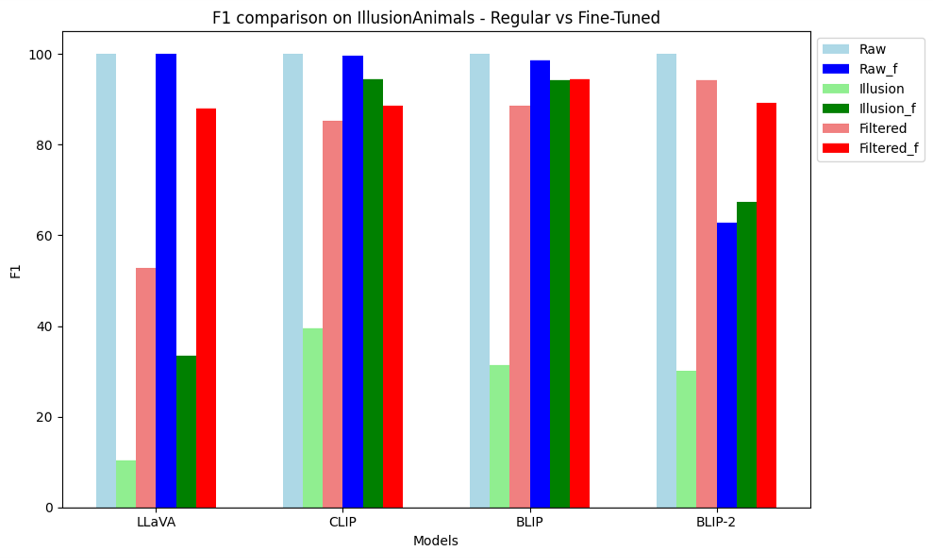}
    \caption{Comparison of zero-shot vs. fine-tuned classification f1 scores on IllusionAnimals dataset}
    \label{fig:f1_illusionanimals}
\end{figure}

You can view the visualizations of the results for both zero-shot and fine-tuned models below (Figures \ref{fig:z_accuracy_comparison}, \ref{fig:z_precision_comparison}, \ref{fig:z_recall_comparison}, \ref{fig:z_f1_comparison} display zero-shot results and figures \ref{fig:accuracy_comparison}, \ref{fig:precision_comparison}, \ref{fig:recall_comparison}, \ref{fig:f1_comparison} display fine-tuned version). You can also find a comparison of the F1 scores for zero-shot and fine-tuned models in the figures \ref{fig:f1_illusionmnist}, \ref{fig:f1_illusionfashionmnist}, \ref{fig:f1_illusionanimals}.

\section{Confusion Matrices}
\label{appendix:confusion_matrices}

In this section, we present the confusion matrices for the responses generated by GPT-4o and the overall human evaluations across three different datasets: IllusionAnimals, IllusionMNIST, and IllusionFashionMNIST.

\subsection{Confusion Matrices for GPT-4o}

\begin{figure}[h]
    \centering
    \includegraphics[width=0.9\columnwidth]{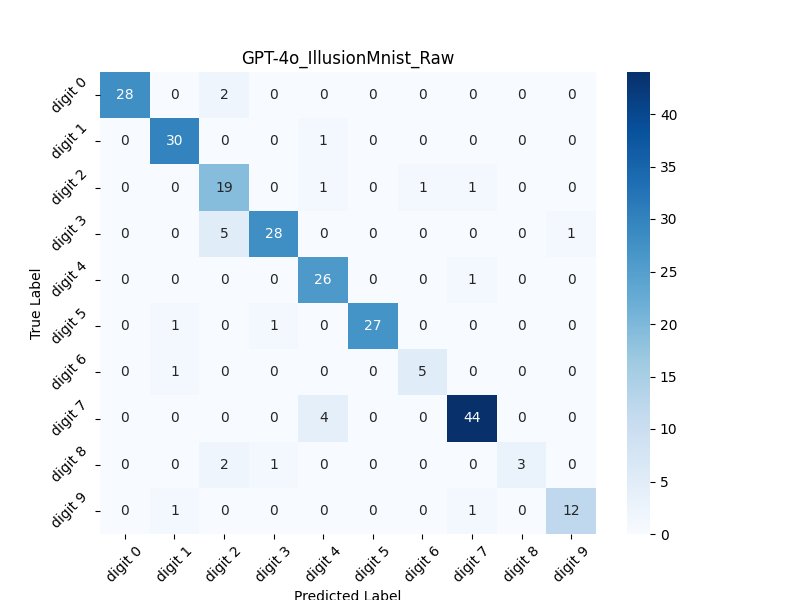}
    \caption{Confusion Matrix for GPT-4o on IllusionMNIST (Raw)}
    \label{fig:confusion_gpt4o_ill_mnist_raw}
\end{figure}

\begin{figure}[h]
    \centering
    \includegraphics[width=0.9\columnwidth]{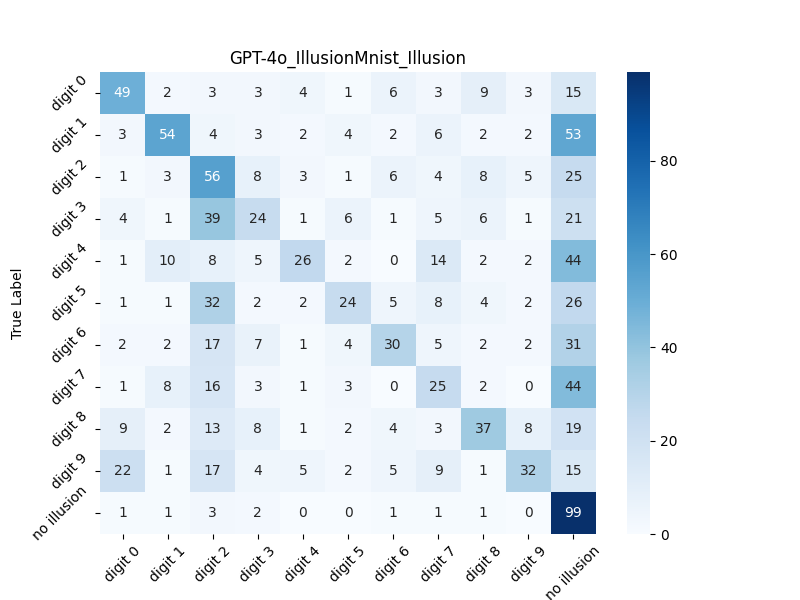}
    \caption{Confusion Matrix for GPT-4o on IllusionMNIST (Illusion)}
    \label{fig:confusion_gpt4o_ill_mnist_ill}
\end{figure}

\begin{figure}[h]
    \centering
    \includegraphics[width=0.9\columnwidth]{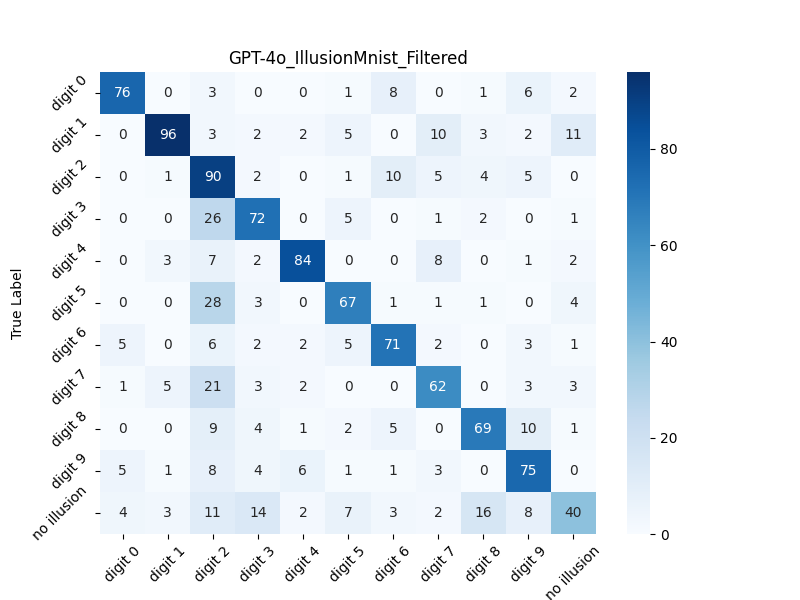}
    \caption{Confusion Matrix for GPT-4o on IllusionMNIST (Filtered)}
    \label{fig:confusion_gpt4o_ill_mnist_filtered}
\end{figure}

\begin{figure}[h]
    \centering
    \includegraphics[width=0.9\columnwidth]{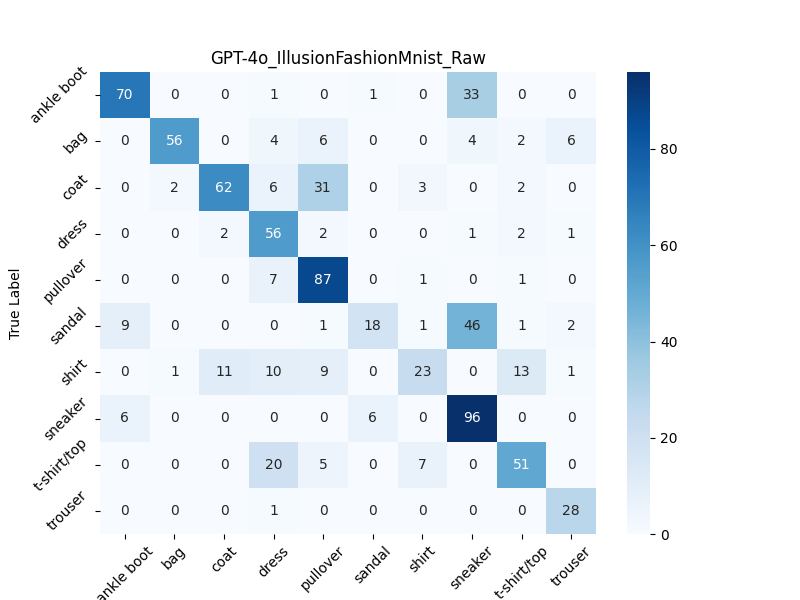}
    \caption{Confusion Matrix for GPT-4o on IllusionFashionMNIST (Raw)}
    \label{fig:confusion_gpt4o_ill_fashionmnist_raw}
\end{figure}

\begin{figure}[h]
    \centering
    \includegraphics[width=0.9\columnwidth]{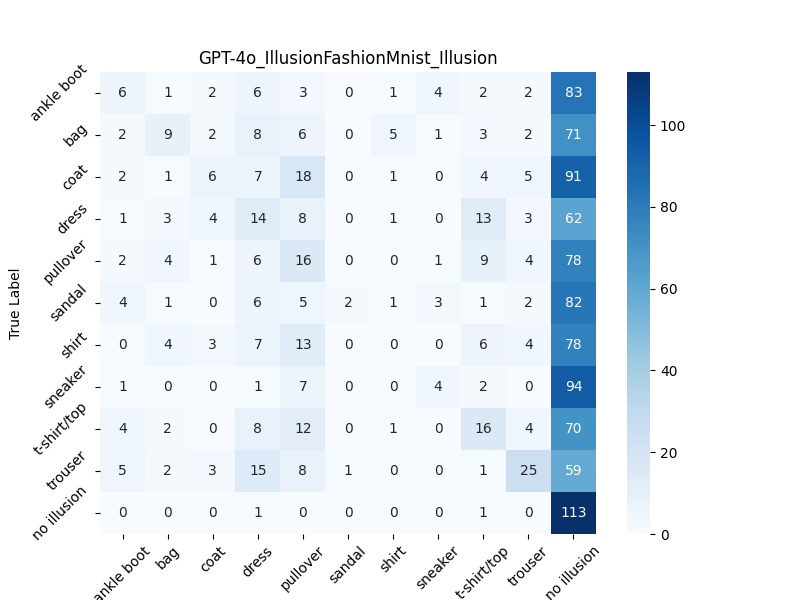}
    \caption{Confusion Matrix for GPT-4o on IllusionFashionMNIST (Illusion)}
    \label{fig:confusion_gpt4o_ill_fashionmnist_ill}
\end{figure}

\begin{figure}[h]
    \centering
    \includegraphics[width=0.9\columnwidth]{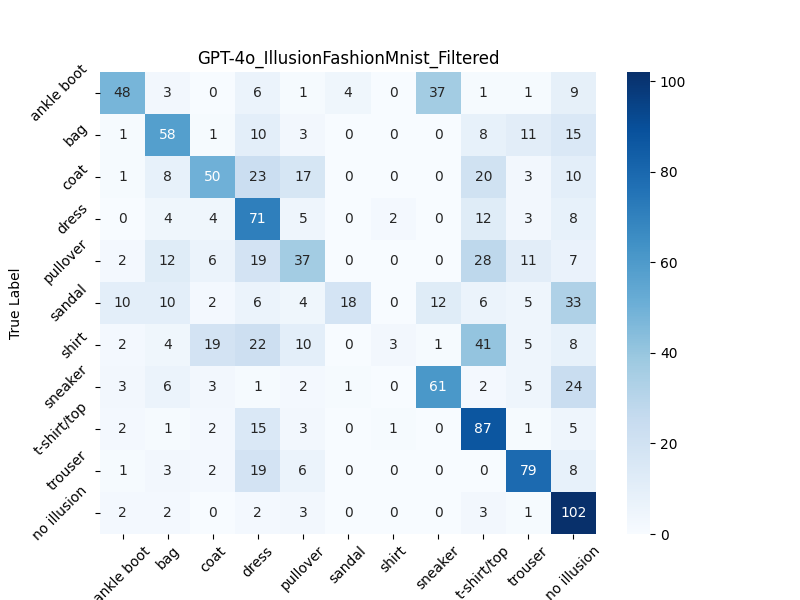}
    \caption{Confusion Matrix for GPT-4o on IllusionFashionMNIST (Filtered)}
    \label{fig:confusion_gpt4o_ill_fashionmnist_filtered}
\end{figure}

\begin{figure}[h]
    \centering
    \includegraphics[width=0.9\columnwidth]{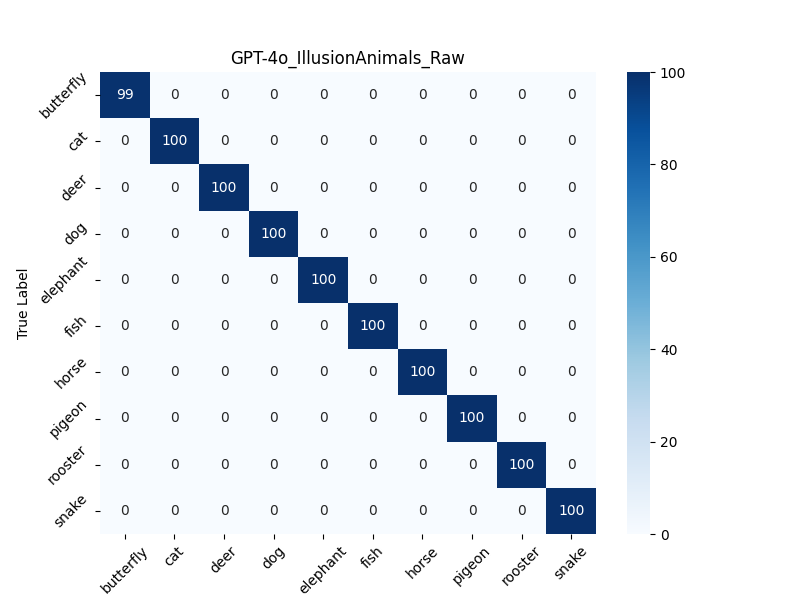}
    \caption{Confusion Matrix for GPT-4o on IllusionAnimals (Raw)}
    \label{fig:confusion_gpt4o_ill_animal_raw}
\end{figure}

\begin{figure}[h]
    \centering
    \includegraphics[width=0.9\columnwidth]{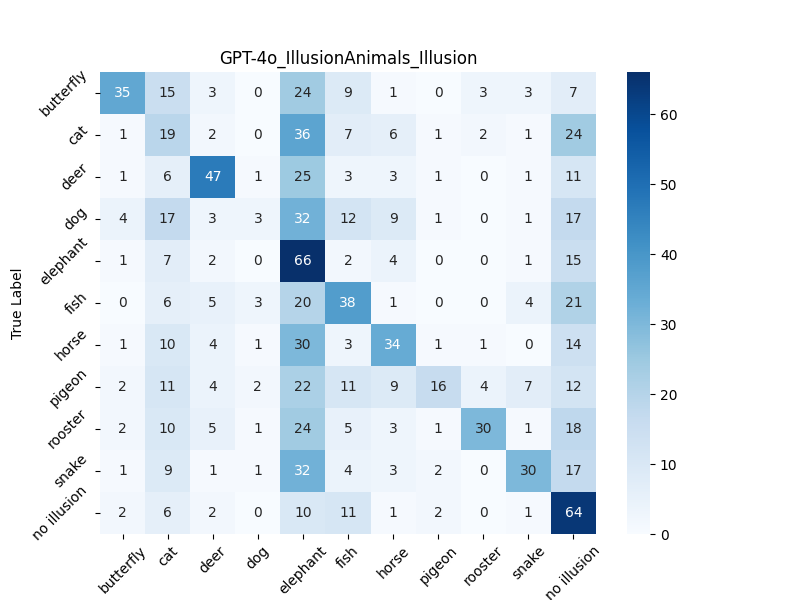}
    \caption{Confusion Matrix for GPT-4o on IllusionAnimals (Illusion)}
    \label{fig:confusion_gpt4o_ill_animal_ill}
\end{figure}

\begin{figure}[h]
    \centering
    \includegraphics[width=0.9\columnwidth]{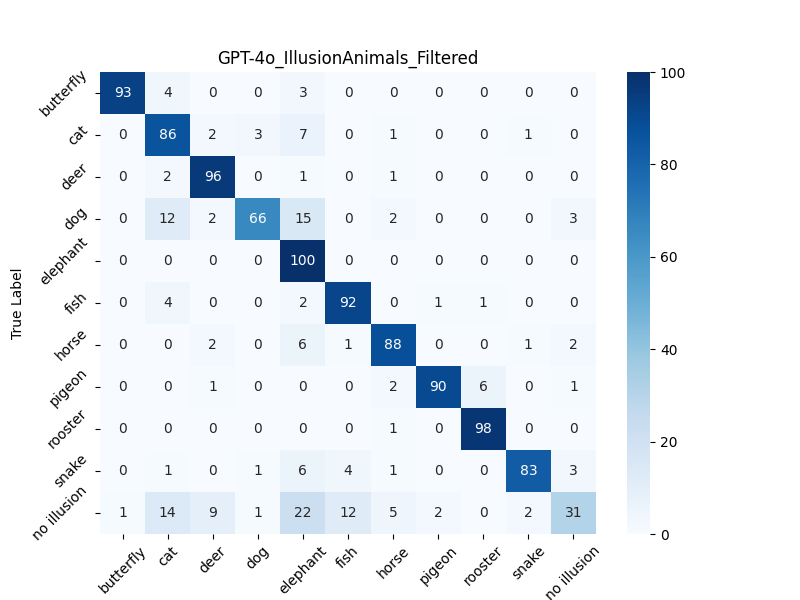}
    \caption{Confusion Matrix for GPT-4o on IllusionAnimals (Filtered)}
    \label{fig:confusion_gpt4o_ill_animal_filtered}
\end{figure}

Figures \ref{fig:confusion_gpt4o_ill_mnist_raw}, \ref{fig:confusion_gpt4o_ill_mnist_ill}, \ref{fig:confusion_gpt4o_ill_mnist_filtered}, \ref{fig:confusion_gpt4o_ill_fashionmnist_raw}, \ref{fig:confusion_gpt4o_ill_fashionmnist_ill}, \ref{fig:confusion_gpt4o_ill_fashionmnist_filtered}, \ref{fig:confusion_gpt4o_ill_animal_raw}, \ref{fig:confusion_gpt4o_ill_animal_ill}, \ref{fig:confusion_gpt4o_ill_animal_filtered} display the confusion matrices for the answers provided by GPT-4o on each part (Raw, Illusion, Filtered) of each dataset. We observe that GPT-4o demonstrates reasonably good performance on raw images across all classes. However, when applying illusions to the images, the values on the main diagonal of the confusion matrix decrease, indicating that it is challenging for GPT-4o to detect illusions. Conversely, after applying our filter to the illusory images, the values on the main diagonal of the confusion matrix increase, highlighting the effectiveness of our method in detecting illusions in images.

\subsection{Confusion Matrices for Human Evaluations}

\begin{figure}[h]
    \centering
    \includegraphics[width=0.9\columnwidth]{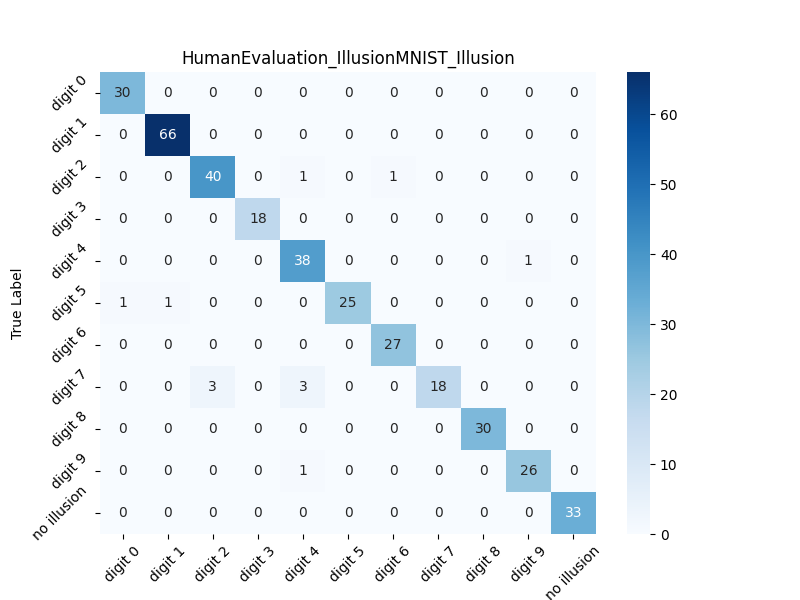}
    \caption{Confusion Matrix for Human Evaluation on IllusionMNIST (Illusion)}
    \label{fig:confusion_human_ill_mnist_ill}
\end{figure}

\begin{figure}[h]
    \centering
    \includegraphics[width=0.9\columnwidth]{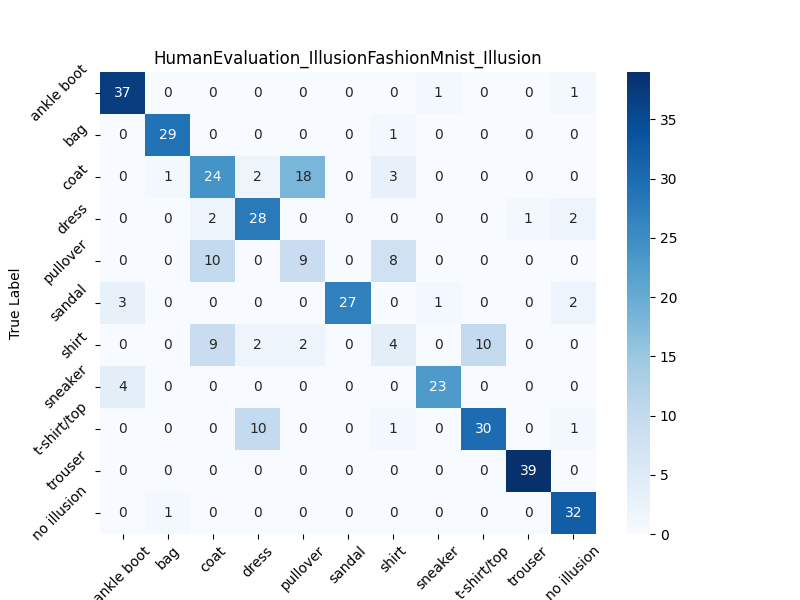}
    \caption{Confusion Matrix for Human Evaluation on IllusionFashionMNIST (Illusion)}
    \label{fig:confusion_human_ill_fashionmnist_ill}
\end{figure}

\begin{figure}[h]
    \centering
    \includegraphics[width=0.9\columnwidth]{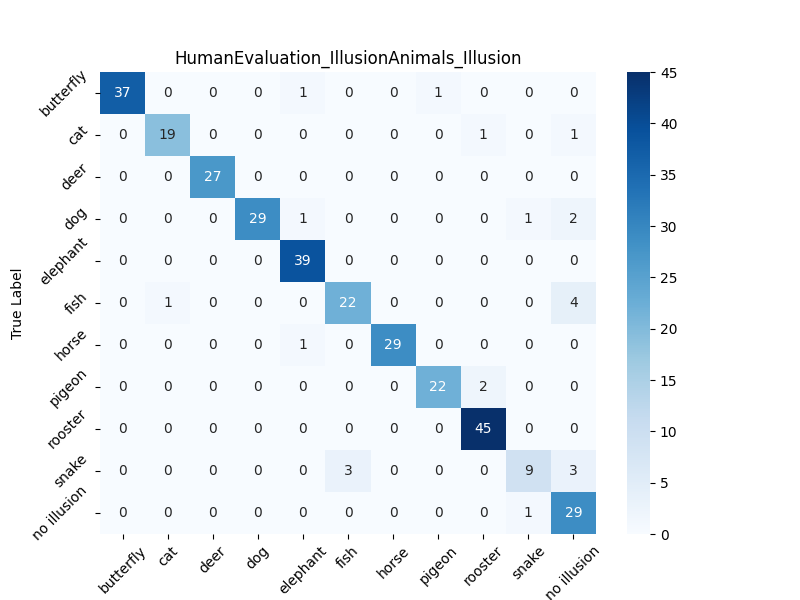}
    \caption{Confusion Matrix for Human Evaluation on IllusionAnimals (Illusion)}
    \label{fig:confusion_human_ill_animals_ill}
\end{figure}

Figures \ref{fig:confusion_human_ill_mnist_ill}, \ref{fig:confusion_human_ill_fashionmnist_ill}, \ref{fig:confusion_human_ill_animals_ill} display the confusion matrices for the overall human evaluations on the Illusion part of each dataset. These matrices show the distribution of predictions made by human annotators for each true label.

\section{Prompt Templates for GPT-4o}
We use the following instructions to prompt GPT-4o.

For raw samples of IllusionMNIST, IllusionFashionMNIST, and IllusionAnimals datasets, we use the following prompt template:

\textit{\begin{quote}
``Which class is in the picture: \{raw\_class\_names\_str\}. Just choose the correct class without any extra explanation.''
\end{quote}}

The raw class names are from the following labels:
\begin{itemize}
    \item \textbf{MNIST}: digit 0, digit 1, digit 2, digit 3, digit 4, digit 5, digit 6, digit 7, digit 8, digit 9
    \item \textbf{Fashion-MNIST}: t-shirt/top, trouser, pullover, dress, coat, sandal, shirt, sneaker, bag, ankle boot
    \item \textbf{Animals}: cat, dog, pigeon, butterfly, elephant, horse, deer, snake, fish, rooster
\end{itemize}

For raw samples of the IllusionChar dataset, we use the following template:
\textit{\begin{quote}
``What sequence of characters are in the picture? Just say the sequence. Put your answer in quotation marks.''
\end{quote}}

For illusion and filtered samples of IllusionMNIST, IllusionFashionMNIST, and IllusionAnimals datasets, we use the following template:
\textit{\begin{quote}
``There might be an illusion of something in the image or not. These are the classes that an illusion might belong to: \{illusion\_class\_names\_str\}. Just choose the correct class without any extra explanation.''
\end{quote}}

The illusion and filtered class names are the same as the raw class names with an additional "No illusion" class.

For illusion and filtered samples of the IllusionChar dataset, we use the following template:
\textit{\begin{quote}
``There might be an illusion of a sequence of characters in the picture. If you cannot detect the sequence of characters, answer with "No illusion". If you can detect the sequence of characters, what sequence of characters are in the picture? Just say the sequence. Put your answer in quotation marks.''
\end{quote}}

\section{GPT-4o Failed Examples}
\label{appendix:gpt4o_failed_examples}
\begin{figure*}[]
    \centering
    \includegraphics[width=\textwidth]{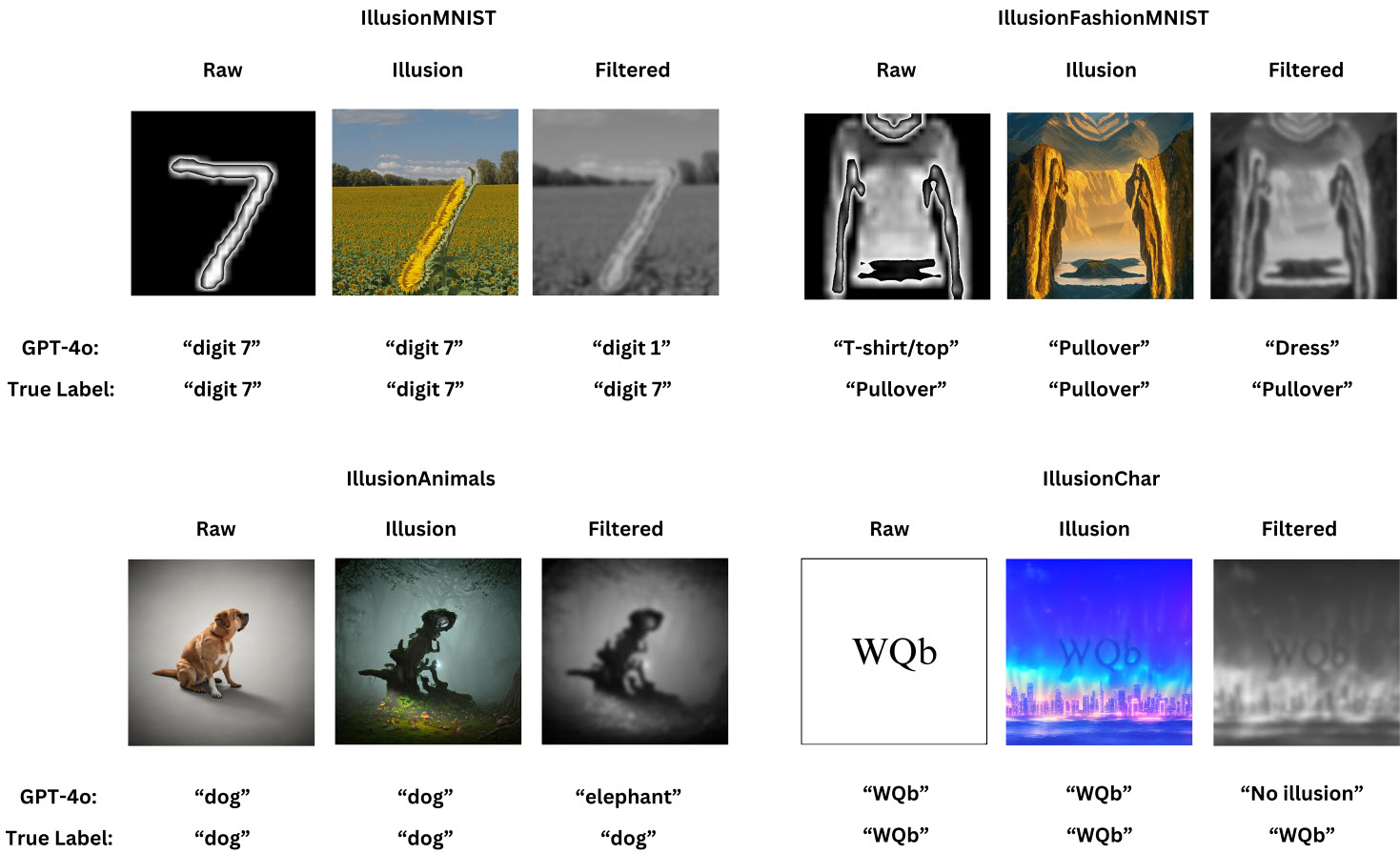}
    \caption{Examples of GPT-4o failing to correctly answer filtered images while correctly answering illusory images.}
    \label{appendix:gpt4o_failure}
\end{figure*}

In this section, we present examples where GPT-4o correctly answered the Illusion image but failed to answer the Filtered image correctly. These examples are taken from four datasets: IllusionAnimals, IllusionMNIST, IllusionFashionMNIST, and IllusionChar. Examples are represented in the Figure~\ref{appendix:gpt4o_failure} 

\section{Hyperparameters}
\label{appendix:hyperparameters}
The detailed hyperparameters for each model and dataset combination are summarized in Tables~\ref{tab:hyperparameters1}, \ref{tab:hyperparameters2}. 

\section{Experimental Setup}
For fine-tuning and inference of the CLIP, BLIP, and BLIP2 models, as well as for inference of the MiniGPT-V2 model on three datasets: IllusionMNIST, IllusionFashionMNIST, and IllusionAnimals, we use the Google Colab T4 GPU. For fine-tuning and inference of the LLaVA model, we use two T4 GPUs on Kaggle. For the Gemini and GPT-4o models, we use the official corresponding APIs.

\begin{table*}[]
\caption{Hyperparameters used for fine-tuning BLIP, BLIP-2, and CLIP Models on three datasets. All these experiments were conducted without a scheduler, using a train/test split of 90\% and 10\%, respectively, and a global seed of 10 to ensure reproducibility.}
\label{tab:hyperparameters1}
\centering
\resizebox{0.7\linewidth}{!}{%
\begin{tabular}{c|c|ccc|}
\cline{2-5}
 & \textbf{Dataset} & \textbf{BLIP} & \textbf{BLIP-2} & \textbf{CLIP} \\ \hline
\multicolumn{1}{|c|}{\multirow{3}{*}{\textbf{Batch Size}}} & \textbf{IllusionMNIST} & 6 & 11 & 6 \\
\multicolumn{1}{|c|}{} & \textbf{IllusionFashionMNIST} & 6 & 11 & 6 \\
\multicolumn{1}{|c|}{} & \textbf{IllusionAnimals} & 6 & 11 & 6 \\ \hline
\multicolumn{1}{|c|}{\multirow{3}{*}{\textbf{Learning Rate}}} & \textbf{IllusionMNIST} & 1e-4 & 1e-5 & 1e-5 \\
\multicolumn{1}{|c|}{} & \textbf{IllusionFashionMNIST} & 5e-5 & 7e-5 & 1e-5 \\
\multicolumn{1}{|c|}{} & \textbf{IllusionAnimals} & 5e-5 & 1e-5 & 1e-5 \\ \hline
\multicolumn{1}{|c|}{\multirow{3}{*}{\textbf{Weight Decay}}} & \textbf{IllusionMNIST} & 1e-4 & 1e-5 & 1e-5 \\
\multicolumn{1}{|c|}{} & \textbf{IllusionFashionMNIST} & 1e-4 & 1e-5 & 1e-5 \\
\multicolumn{1}{|c|}{} & \textbf{IllusionAnimals} & 1e-4 & 1e-5 & 1e-5 \\ \hline
\multicolumn{1}{|c|}{\multirow{3}{*}{\textbf{\# of Epochs}}} & \textbf{IllusionMNIST} & 3 & 3 & 3 \\
\multicolumn{1}{|c|}{} & \textbf{IllusionFashionMNIST} & 3 & 3 & 3 \\
\multicolumn{1}{|c|}{} & \textbf{IllusionAnimals} & 3 & 3 & 3 \\ \hline
\multicolumn{1}{|c|}{\multirow{3}{*}{\textbf{Optimizer}}} & \textbf{IllusionMNIST} & AdamW & AdamW & AdamW \\
\multicolumn{1}{|c|}{} & \textbf{IllusionFashionMNIST} & AdamW & AdamW & AdamW \\
\multicolumn{1}{|c|}{} & \textbf{IllusionAnimals} & AdamW & AdamW & AdamW \\ \hline
\end{tabular}
}
\end{table*}

\begin{table*}[]
\caption{Hyperparameters used for fine-tuning LLaVA model on three datasets, namely IllusionMNIST, IllusionFashionMNIST, and IllusionAnimals}
\label{tab:hyperparameters2}
\centering
\resizebox{0.5\linewidth}{!}{%
\begin{tabular}{|cc|c|}
\hline
\multicolumn{2}{|c|}{\textbf{Hyperparameter}} & \textbf{Value} \\ \hline
\multicolumn{2}{|c|}{\textbf{Learning Rate}} & 1e-5 \\ \hline
\multicolumn{2}{|c|}{\textbf{Batch Size (per device)}} & 8 \\ \hline
\multicolumn{2}{|c|}{\textbf{\# of Epochs}} & 2 \\ \hline
\multicolumn{2}{|c|}{\textbf{Optimizer}} & AdamW \\ \hline
\multicolumn{1}{|c|}{\multirow{4}{*}{\textbf{LoRA}}} & \textbf{r} & 64 \\
\multicolumn{1}{|c|}{} & \textbf{lora\_alpha} & 16 \\
\multicolumn{1}{|c|}{} & \textbf{target\_modules} & all-linear \\
\multicolumn{1}{|c|}{} & \textbf{dropout} & 0 \\ \hline
\end{tabular}
}
\end{table*}

\section{Filter Details}
\label{appendix: filter}
Below, you can find the details of the filter we used to apply to illusory images for detecting illusions. We developed this filter, drawing inspiration from the human eye's ability to detect visual illusions when partially closed.

\begin{lstlisting}[language=Python]
import cv2
from google.colab.patches import cv2_imshow
import numpy as np

def generate_filtered_image(image_path, blur_amount=61):
    image = cv2.imread(f"{image_path}")
    blurred_image = cv2.GaussianBlur(image, (blur_amount, blur_amount), 0)
    blurred_image = cv2.blur(blurred_image, (20, 20))
    blurred_image = cv2.medianBlur(blurred_image, 5)
    gray_image = cv2.cvtColor(blurred_image, cv2.COLOR_BGR2GRAY)
    kernel = np.array([[-1, -2, -1], [-2, 13, -2], [-1, -2, -1]])
    sharpened_image = cv2.filter2D(gray_image, -1, kernel)
    return sharpened_image
\end{lstlisting}

\end{document}